\documentclass[sigconf]{acmart}

\usepackage{algorithm}
\usepackage{algorithmic}

\AtBeginDocument{%
  \providecommand\BibTeX{{%
    \normalfont B\kern-0.5em{\scshape i\kern-0.25em b}\kern-0.8em\TeX}}}

\setcopyright{acmcopyright}
\copyrightyear{2023}
\acmYear{2023}

\acmJournal{TOG}
\copyrightyear{2023} 
\acmYear{2023} 
\setcopyright{rightsretained} 
\acmConference[SA Conference Papers '23]{SIGGRAPH Asia 2023 Conference Papers}{December 12--15, 2023}{Sydney, NSW, Australia}
\acmBooktitle{SIGGRAPH Asia 2023 Conference Papers (SA Conference Papers '23), December 12--15, 2023, Sydney, NSW, Australia}
\acmDOI{10.1145/3610548.3618230}
\acmISBN{979-8-4007-0315-7/23/12}

\begin{document}

\title{Animating Street View}

\author{Mengyi Shan}
\email{shanmy@cs.washington.edu}

\affiliation{%
  \institution{University of Washington}
  \city{Seattle}
  \state{WA}
  \country{US}
}
\author{Brian Curless}
\email{curless@cs.washington.edu}

\affiliation{%
  \institution{University of Washington}
  \city{Seattle}
  \state{WA}
  \country{US}
}

\author{Ira Kemelmacher-Shlizerman}
\email{kemelmi@cs.washington.edu}

\affiliation{%
  \institution{University of Washington}
  \city{Seattle}
  \state{WA}
  \country{US}
}

\author{Steve Seitz}
\email{seitz@cs.washington.edu}

\affiliation{%
  \institution{University of Washington}
  \city{Seattle}
  \state{WA}
  \country{US}
}

\renewcommand{\shortauthors}{Shan, et al.}

\begin{abstract}
We present a system that automatically brings street view imagery to life by populating it with naturally behaving, animated pedestrians and vehicles. Our approach is to remove existing people and vehicles from the input image, insert moving objects with proper scale, angle, motion and appearance, plan paths and traffic behavior, as well as render the scene with plausible occlusion and shadowing effects. The system achieves these by reconstructing the still image street scene, simulating crowd behavior, and rendering with consistent lighting, visibility, occlusions, and shadows. We demonstrate results on a diverse range of street scenes including regular still images and panoramas.

\end{abstract}



\begin{teaserfigure}
  \includegraphics[width=\textwidth]{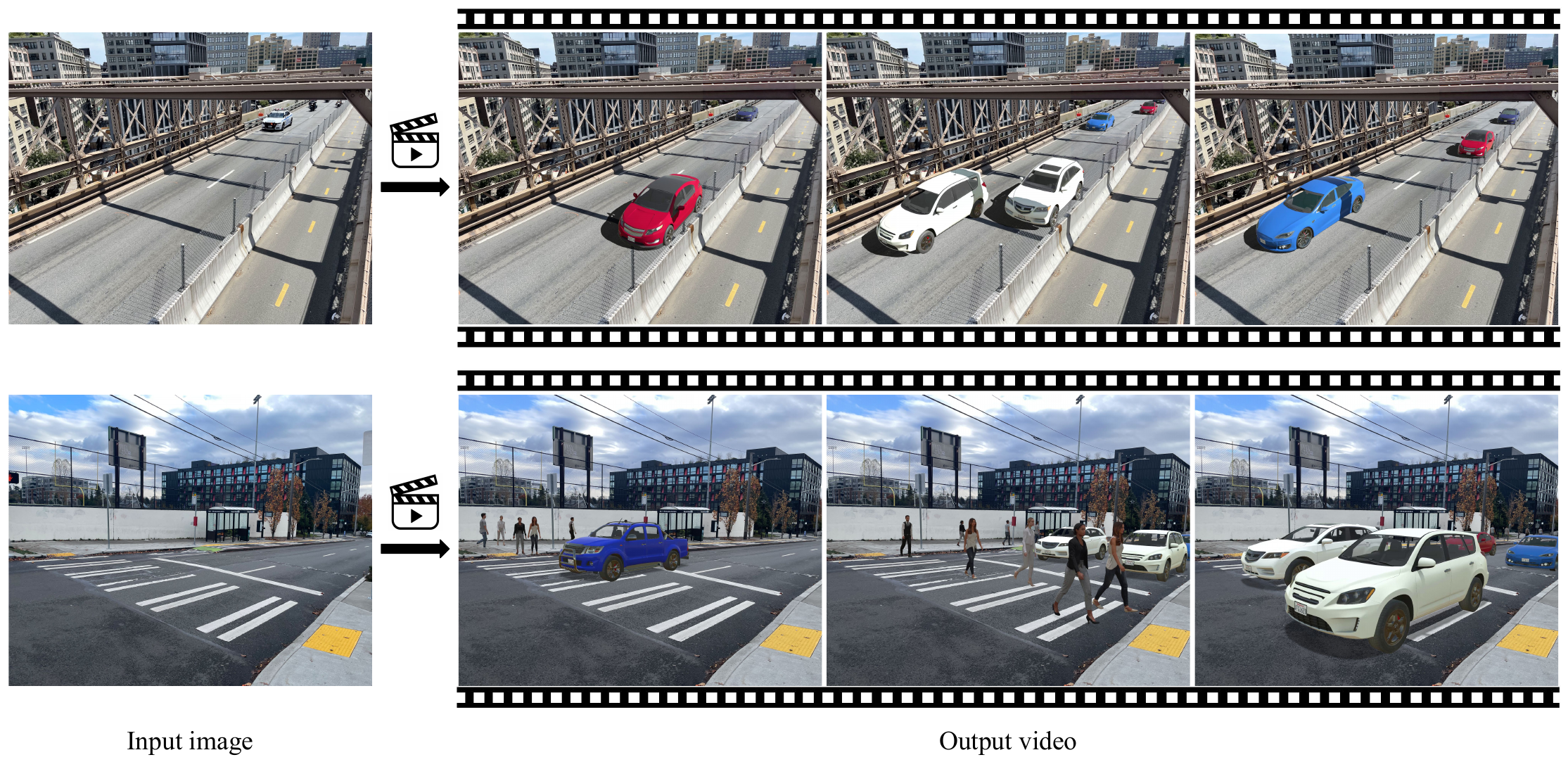}
  \caption{Given a single input street image or panorama, our system automatically removes existing people or vehicles, and populates it with simulated animated pedestrians and cars. The result is a high resolution, arbitrarily long video. Our system handles scaling,  placement, path planning and traffic simulation, as well as estimates and renders correct lighting, occlusion, and shadows.}
  \label{fig:teaser}
\end{teaserfigure}


\maketitle

\setlength{\textfloatsep}{5pt}
\section{Introduction}
Google Street View and similar services allow people to virtually visit  locations worldwide with street-level imagery. These platforms offer an immersive experience of different areas such as neighborhoods, streets, and tourist attractions. Nevertheless, the major limitation of the visual content provided by these services is that they are all still -- notably, imagery without people or cars moving through each scene.  Street-level video could address this shortcoming but would be difficult due to constraints on image capture systems, storage, and privacy requirements. We propose a system that mitigates these limitations by first erasing still pedestrians and vehicles from an image or panorama of a scene and repopulating it with synthetic, moving pedestrians and vehicles. Our method can thus enhance the vividness of street view imagery without additional capture and without privacy concerns. 

\begin{figure*}[h]
  \centering
  \includegraphics[width=\linewidth]{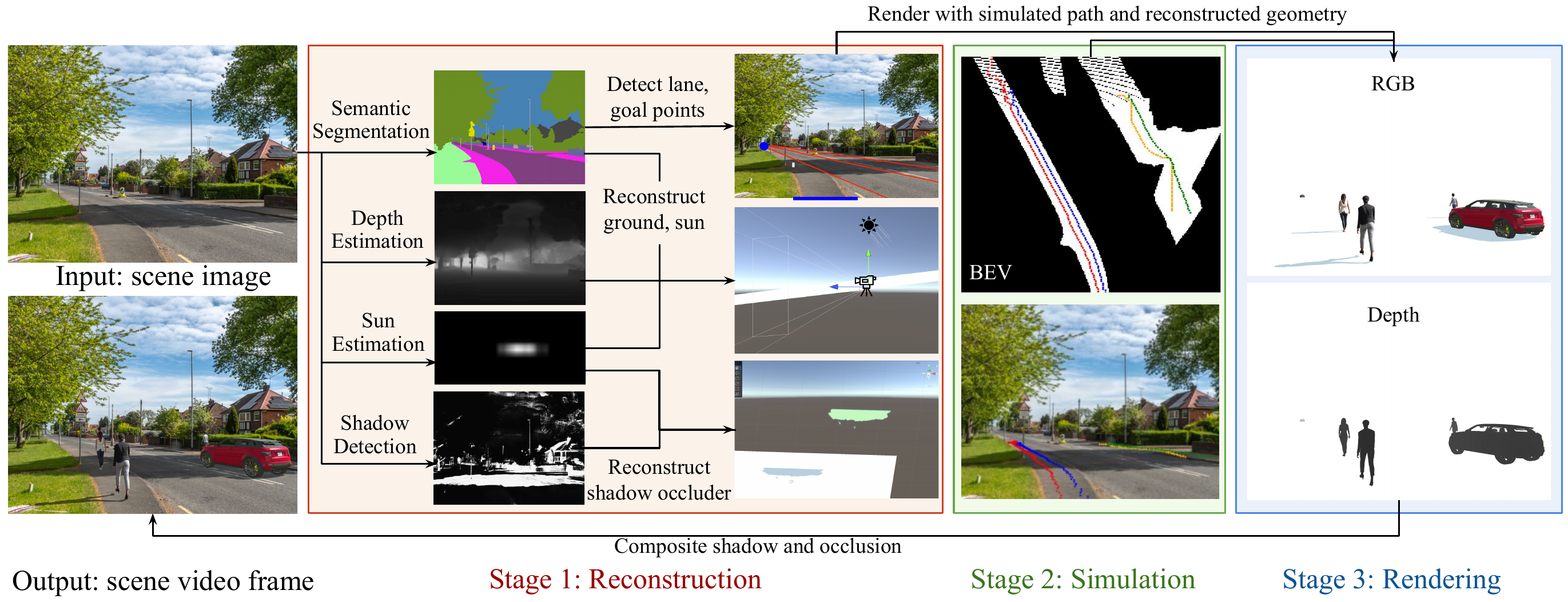}
  \caption{Our system has three major components. In Stage 1, we reason about the scene by predicting its semantic segmentation labels, depth values, sun direction and intensity, as well as shadow regions. We additionally determine walking and driving regions for adding pedestrians and cars (red straight lines: lane detection; blue points: origin and destination points). In Stage~2, we simulate the pedestrians in a 2D bird's eye view representation (BEV) of the scene, and simulate car movements with predicted lanes (four colors correspond to four predicted path, both in BEV and scene images). If there is a detected crosswalk, we also simulate the traffic behavior by controlling a traffic light (not shown in this example, refer to Figure \ref{fig:traffic}). In Stage 3, we render the scene with the estimated lighting, shadows, and occlusions. The whole pipeline is automated. }
  \label{fig:pipeline}
\end{figure*}

Our framework takes as input a single street image or panorama and generates a video of arbitrary length by populating it with moving pedestrians and vehicles. In order for the generated videos to look realistic, they must respect the geometry, lighting, and semantic information -- e.g., labeling sidewalks and streets -- of the given scene, while modeling reasonable behaviors of newly added elements (people and vehicles).  Our method takes all these factors into consideration in three main stages (as shown in Figure \ref{fig:pipeline}):

\begin{enumerate}
    \item \textbf{Reconstruction}: estimating the visible geometry, lighting, and semantic information of a scene. 
    \item \textbf{Simulation}: inserting and modeling the behavior of pedestrians and vehicles in a given street setting.
    \item \textbf{Rendering}: synthesizing realistic videos with consistent lighting, shadows, and occlusions layered into the original frames.

\end{enumerate}

Our method is the first to demonstrate realistic results for street scene image/panorama animation, producing arbitrarily long videos with moving people and cars. We combine learning-based approaches with traditional rendering to deliver visually appealing results without large training sets, large scale annotations, long training time or advanced hardware. 
Our approach leverages a number of ingredients from prior art.
Specifically, our geometry \textit{reconstruction}, \textit{crowd simulation} and shadow/occlusion \textit{rendering} components build on existing pre-trained networks and rendering systems, except for a sun estimation sub-module that we trained. And while these ingredients by themselves are not novel, our system combines them in novel ways to enable an exciting new application.
We only require a single image as input with a known focal length, or a 360$^\circ$ panorama. The whole pipeline can be run on a conventional laptop. Figure \ref{fig:teaser} illustrates several output frames for three different scenes. Please refer to the supplementary material for video results.


\section{Related Work}
\label{sec:related}

\subsection{Object Insertion} 

 Early insertion work includes Poisson blending \cite{Prez2003PoissonIE} which produces seamless object boundaries but results in illumination and color mismatches between object and the target background. \cite{karsch2014insertion} and \cite{Karsch2011insertion} recover geometry and lighting from a single image for object insertion. However, \cite{Karsch2011insertion} requires manual insertion, and neither demonstrates automatic occlusion behavior, shadow cast by the scene onto objects, or object placement/movement. More recent work renders inserted objects with estimated lighting maps and shadow synthesis \cite{wang2022neural, tang2022lighting}, but does not handle sharp lighting changes near shadow boundaries. A number of recent methods learn end-to-end object insertion and composition based on GANs \cite{zhan2019sfgan, lin2018stgan, azadi2018compositional} and diffusion models \cite{ma2023directed, song2022objectstitch}. 

Specifically for street scenes, \cite{chien2017nonexistent, lee2018insert, hiddenfootprints2020eccv} estimate the locations of roads and walking paths. \cite{wang20people, wang2021repopulating} {\em manually} insert {\em static} pedestrians and vehicles into a scene image, selected to have scene-compatible lighting and pose, but do not support animation. \cite{wang20people} also requires video as input. In contrast, our approach inserts 3D dynamical subjects, is automatic, and operates on a single image.

The problem of animating street scenes has been less well studied.
\cite{Lee2019InsertingVI} learn to composite a given video clip of a pedestrian onto a scene video, but with limited quality. \cite{xu2022discoscene} train a 3D-aware GAN given scene image, but only show short (1-2 seconds) video clips with noticeable artifacts. Closest to our work is \cite{chen2021geosim} which composes moving car assets into a scene video, and \cite{wang2022neural}, but they require 5-7 posed RGB imagery and LiDAR for NeRF or 3D-aware geometry representation. \cite{chen2021geosim} also only supports vehicles instead of the much more complex pedestrian/vehicle interaction. Compared with previous works, our system only requires a still image, and handles pedestrians in addition to vehicles. 

\subsection{Still Image Animation} 
Image animation aims to generate a video given one or a sequence of still images as input. \cite{okabe2011fluid, holynski2021animating} show beautiful animation results assuming repetitive motions, e.g., waterfalls. \cite{yoon2021poseguided, huang2022physically, peng2021animatable, weng2019wakeup}  animate existing people in the scene, and \cite{mallya2022implicit, pumarola2019ganimation} animate faces.  \cite{wang2019fewshotvid2vid, mallya2022implicit} take in conditional signals like facial keypoints or pose sequences provided by a driving video and synthesize video in 2D space, and \cite{huang2022elicit} reconstruct an animatable implicit representation from a still image. \cite{hu2022makeitmove, yu2022semantic} animate a single image by semantic instructions.  There are also a line of work on 3D-aware GAN for scene generation (\cite{niemeyer2020GIRAFFE, xue2022giraffehd, BlockGAN2020, epstein2022blobgan}), which can be leveraged for scene animation by GAN inversion.  Recent methods generate full videos via diffusion models, e.g.,  \cite{ni2023conditional} focuses on human motion and fixed camera pose, and \cite{karras2023dreampose} show animation specific to fashion models. We focus on 3D scene modeling for the purpose of occlusion-aware and lighting-consistent animation of novel objects' interaction within the scene. 



\subsection{Playable Video Generation}
\cite{menapace2021pvg, menapace2022pvg} address Playable Video Generation that learns semantically consistent actions and generates realistic videos conditioned on the input. Similarly, \cite{zhang2021vid2player, zhang2023vid2player3d} produce videos by modeling interactively controllable video sprites and composing them onto an empty scene. \cite{davtyan2022cvg}  learn to segment video into foreground-background layers and generate transitions of the foreground over time. \cite{kim2021drivegan} learn to simulate a dynamic driving environment directly in pixel-space. These methods require a large video dataset for training and are thus limited to a  narrow set of scene images, for example a tennis court from a fixed angle, or a first-person driving scene. Our method is the first approach to animate street scene without heavy training by populating it with naturally behaving characters.

\section{Approach}
\label{sec:method}

Given an image of a street scene, our goal is to generate an arbitrarily long animation of this scene populated with objects including pedestrians and vehicles. The resulting video should respect the geometry and illumination of the scene, contain realistic shadow and occlusion effects, and feature natural traffic behaviors. 

Our automatic pipeline contains three stages. Section \ref{sec:reconstruction} describes a reconstruction stage where the basic geometry and illumination are estimated and reconstructed. Section \ref{sec:simulation} describes a simulation stage where pedestrians' and vehicles' behaviors are simulated in the reconstructed scene. Section \ref{sec:rendering} describes a rendering stage where characters are placed into the scene and processed to generate desired visual effects. 

For simplicity, this section focuses on traditional camera images and omits discussion of hyperparameters. For equirectangular panoramas, we decompose the imagery into six perspective images in a cubemap manner and process each direction separately. A comprehensive discussion of hyperparameters and panorama processing can be found in the supplementary materials. 

\subsection{Reconstruction}
\label{sec:reconstruction}

In this section, we describe the reconstruction process, which removes existing pedestrians and vehicles from the image, and estimates the (1) semantic information to determine walking and driving regions for the scene, (2) ground plane for deciding scale and camera angles, (3) sun light direction, intensity and ambient light intensity for lighting the objects, and (4) existing shadow regions in the image for darkening objects when they move into shadow.

\subsubsection{Inpainting}
If an image contains pedestrians and/or vehicles, we start by removing them with segmentation and inpainting. 

\begin{figure}[h]
  \centering
  \includegraphics[width=\linewidth]{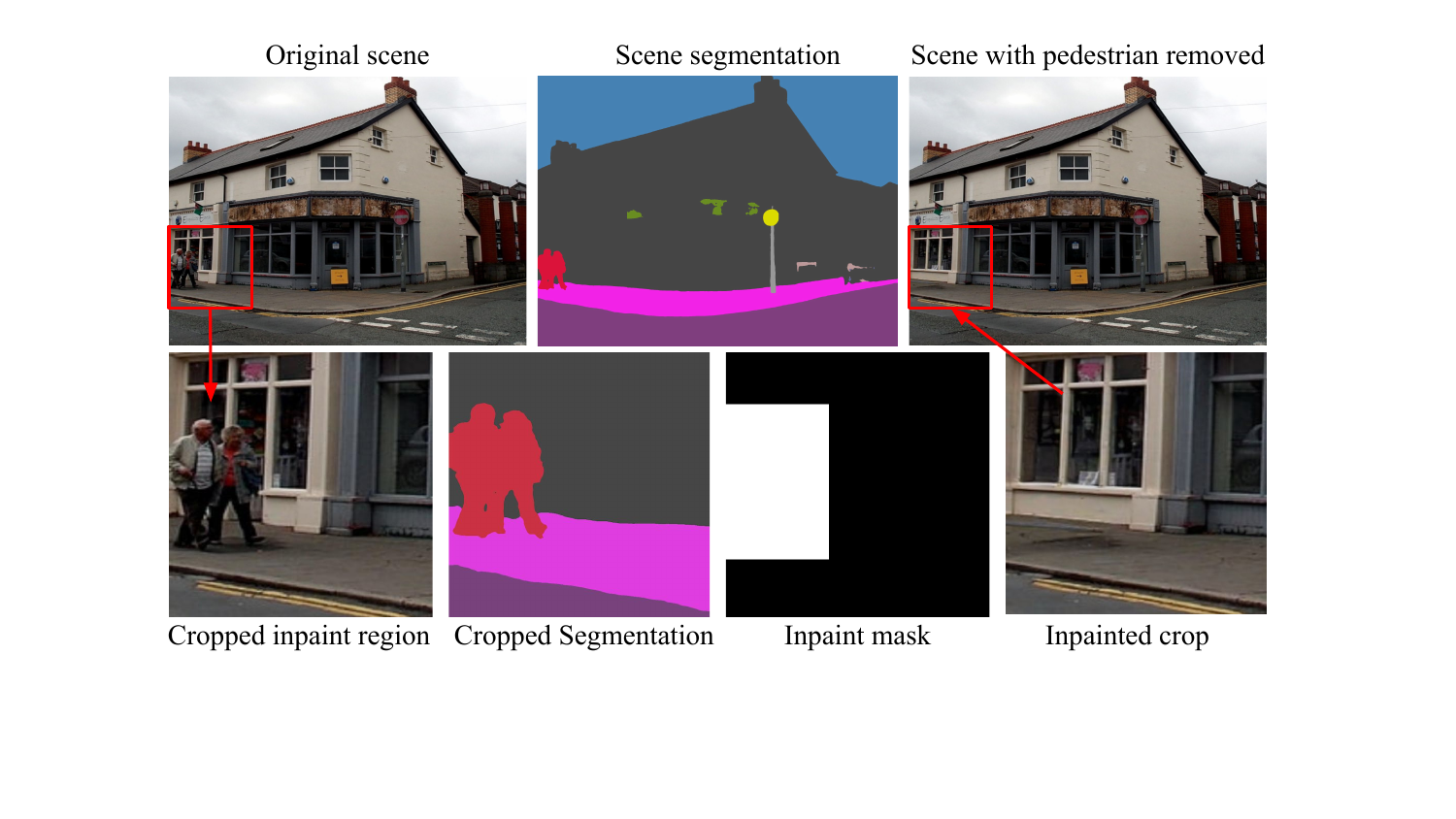}
  \caption{Stable Diffusion \cite{Rombach2022SD} based inpainting. We detect existing people and cars with segmentation map, crop the surrounding region, inpaint and compose back to the image. }
  \label{fig:inpainting}

\end{figure}

As illustrated in Figure \ref{fig:inpainting}, we directly take advantage of the pre-trained Stable Diffusion inpainting tool \cite{Rombach2022SD} to remove existing objects. We segment people and cars using SegFormer \cite{xie2021segformer} and compute a bounding box for inpainted objects as a rectangle covering the objects with a $10\%$ margin to accommodate boundary inaccuracy in semantic segmentation. We use \textit{"A photo of an empty street"} as the inpainting prompt, and \textit{"Human, pedestrian, vehicle, car"} as the negative prompt for Stable Diffusion. 

\subsubsection{Segmentation}

We segment the image to determine where the pedestrians can walk or the vehicles can drive. We start with a semantic segmentation module with off-the-shelf semantic segmentation model SegFormer \cite{xie2021segformer} to segment out the ground regions (\texttt{sidewalk} and \texttt{road} classes in CityScapes \cite{Cordts2016Cityscapes}). However, we found SegFormer \cite{xie2021segformer} sometimes identifies \texttt{Road} region as not suitable for driving especially when the scene is not a traditional CityScapes \cite{Cordts2016Cityscapes} style image. We thus apply an additional zero-shot language-driven segmentation model CLIPSeg \cite{lueddecke2022clipseg} to help us segment the image with keyword "drive". CLIPSeg is adaptable to more segmentation classes in a wider range of scenes, but produces less accurate pixel-level boundaries. If the identified "drive" region is smaller than a threshold $t_\text{d}$, we make the scene "pedestrian-only" by considering the \texttt{Road} label pixels also as \texttt{Sidewalk} pixels, and not adding vehicles to the scene. 

To handle the case where there are obstacles on street that partition the walking/driving regions into pieces -- e.g., the a pole in the image may split the visible sidewalk into disjoint 2D pieces -- we dilate the region and then take the convex hull of each connected component. At the same time, we record the ground position of each obstacle to reconstruct a binary obstacle map with non-walkable obstacle pixels. We utilize the estimated ground plane equation to project the semantic information onto a bird's eye view (BEV) map (Figure \ref{fig:pipeline} top figure in Stage 2), which contains the spatial information for walking/driving regions and obstacle positions. This BEV map is discretized into a grid and serves as the abstract representation for pedestrian simulation. Note that vehicle simulation is a different process and not shown in this BEV map.

\subsubsection{Ground Plane Estimation}

We make the assumption that the ground region in the image can be well-approximated by a plane 
\begin{equation}
    aX + bY + cZ = 1
    \label{eq:plane}
\end{equation} 
\noindent 
and the goal is to estimate the values of $a, b, c$ in metric units (1/meters). This approach is similar to \cite{wang2021repopulating}.

We use the monocular depth estimation model AdaBins \cite{sfaroop2021adabins} to predict the absolute depth map in meters. According to the properties of perspective projection, for known focal length $f$, image pixels $x, y$ and 3D positions $X, Y, Z$, we have: $x = X f / Z$ and $y = Y f / Z$. Plugging into Equation ~\ref{eq:plane} and rearranging yields 
\begin{equation}
    \frac{a}{f} x + \frac{b}{f} y + c = \frac{1}{Z}
\end{equation} 
Given the segmentation map, we collect all 2D road/sidewalk pixels $(x_i, y_i)$ and their depth $Z_i$, and solve for $a, b, c$ via linear least squares. For simplicity, we assume that the road and sidewalk are co-planar.

This plane is then the surface on which pedestrians will walk and vehicles will drive in simulation, augmented with the segmentation map to restrict those regions and the depth map to determine occlusions between the scene and inserted assets.


\subsubsection{Sun Estimation}
\label{sec:sun}
In order to realistically light the inserted objects, we train a sun estimation network that learns to predict the sun direction from a single image. 
We formulate it as a classification problem and predict a distribution over discrete sun angles. We divide the range of azimuth angles $[0,2 \pi)$ into 36 bins and elevation angles $[0,\pi/2)$ into 18 bins, each bin spanning 5 degrees. We train with a ResNet50 \cite{He2015DeepRL}, but replace the last fully connected layer with two, one for azimuth and one for elevation. We train this network using ground truth sun positions as supervision via a cross-entropy loss. This sun estimation approach is closely based on \cite{wang2021repopulating}. We compute the variance of the predicted elevation/azimuth bin distributions to decide if the scene has a strong, directional light or a diffuse environment light, and accordingly set the directional light source strength and ambient light when lighting inserted objects. The thresholds of variance and corresponding light intensities are provided in the supplementary material. 

\begin{figure}[h]
  \centering
  \includegraphics[width=\linewidth]{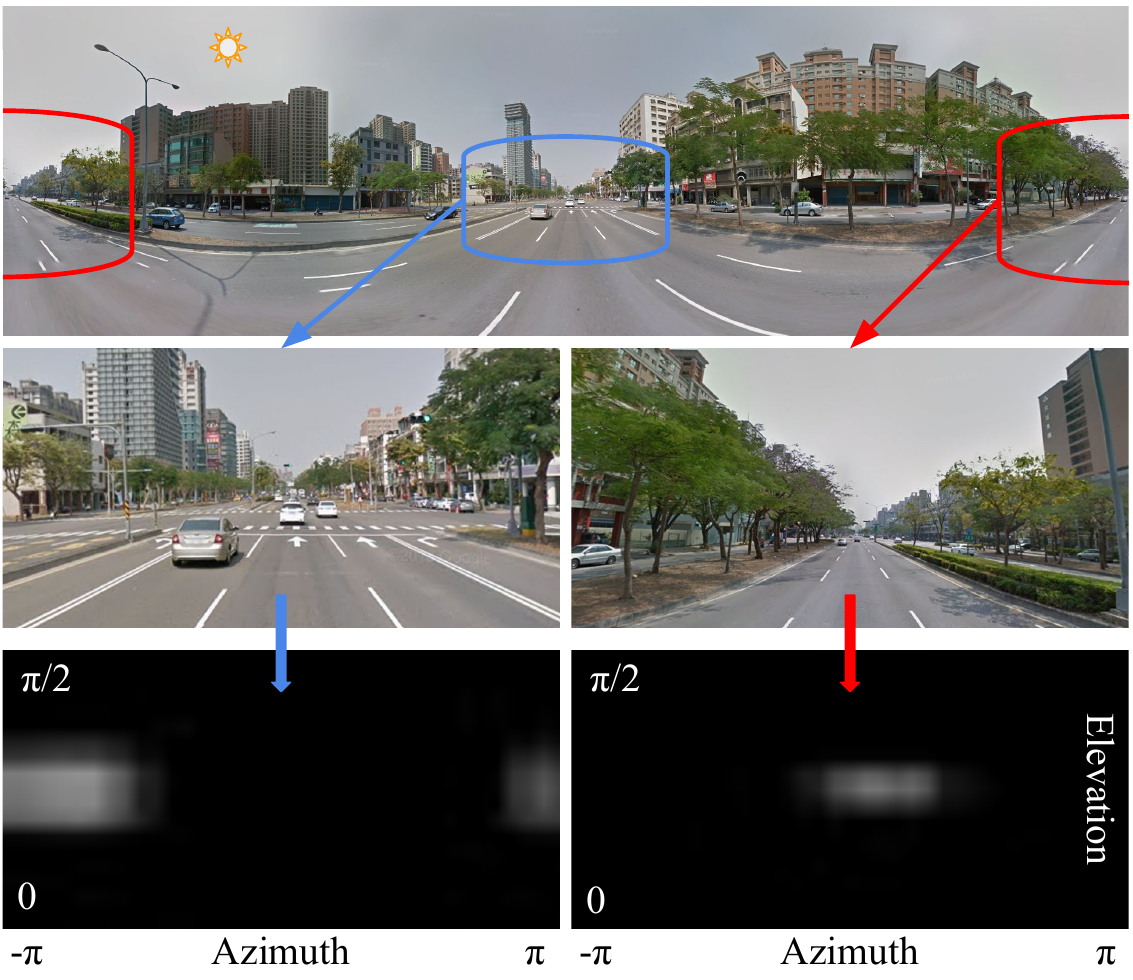}
  \caption{A 360 degree panorama is projected onto planar images with various camera poses and FOVs. Then the system is trained to predict sun light direction (ground truth shown in the panorama) by estimating its elevation and azimuth.  The bottom row visualizes possible light location estimated by the network, shown as the outer product between azimuth and elevation distribution vectors. Note that the sun is in different positions relative to the camera viewpoints for the two crops.}
  \label{fig:sun}
\end{figure}

For training and ground truth data, we use a panorama dataset \cite{chang2018panorama} with estimated sun direction for Google Street View panoramas from sky appearance and metadata. We take perspective images from each panorama with diverse camera angles and FOV, and compute the ground truth sun elevation and azimuth for these images. Refer to Figure \ref{fig:sun} for an example data point, and to Section \ref{sec:result} for a discussion on data preprocessing.

\subsubsection{Scene Shadow Estimation}
Analyzing local shadows in the scene is important for realistic object insertion.  In particular, shadows cast by inserted objects should not darken existing shadows, and shadows cast by elements in the scene, e.g., by in-scene structures or trees, should also be cast onto the inserted objects.

We start by detecting the existing shadows in the image using an off-the-shelf shadow detection network \cite{zhu2021shadow}. We keep ground shadow regions by intersecting this binary map with the ground region obtained from semantic segmentation. 

\begin{figure}[ht!]
  \centering
  \includegraphics[width=\linewidth]{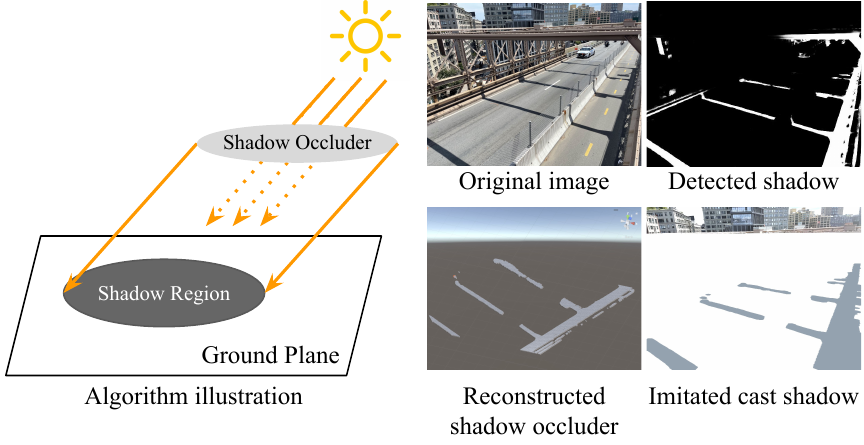}
  \caption{We detect the shadow region in the image, and trace rays in 3D from the corresponding points toward the light source through a fixed plane above all potential objects in the scene. The reconstructed 3D shadow occluder then casts shadows onto objects in the scene.}
  \label{fig:occluder}
\end{figure}

With the estimated ground plane equation and sun direction from previous steps, we now construct a shadow occluder to explain shadows cast into the scene.  In particular, we first create a plane parallel to the ground plane and displaced vertically above the height of any potentially inserted objects and then intersect rays shadow rays -- from the 3D positions of shadowed scene points toward the light direction -- with this plane.  We connect these points into a mesh to construct the shadow occluder and use it to cast shadows onto inserted objects.  See Figure \ref{fig:occluder} for an illustration of this pipeline.


\subsection{Simulation}
\label{sec:simulation}

After reconstructing the basic scene geometry and semantics, we utilize the scene understanding information to simulate the movement of pedestrians and vehicles in the scene. We start with estimating origin and destination points for pedestrian movements, and then run a simple yet effectively potential field-based crowd simulation algorithm. For scenes with more complex structure, we also design a traffic simulation module that naturally controls the behavior of pedestrians and cars at traffic lights and crosswalks. 

\subsubsection{Origin/Destination Estimation}
We would like each pedestrian to either appear or disappear from one edge of the image and disappear or appear, respectively, at the far end of the scene. Thus we build a pool of possible origin/destination pairs containing the intersection of walkable regions and image boundaries, as well as the farthest points from the camera. To simulate each pedestrian, we randomly select one pair of origin and destination from this pool and check if there is an existing pedestrian in this same or neighboring grid. If not, we initialize a pedestrian in this grid. 

For cars, we split the drivable region into lanes by estimating the width of the drivable region. See more details in supplementary. 

\begin{figure}[h]
  \centering
  \includegraphics[width=\linewidth]{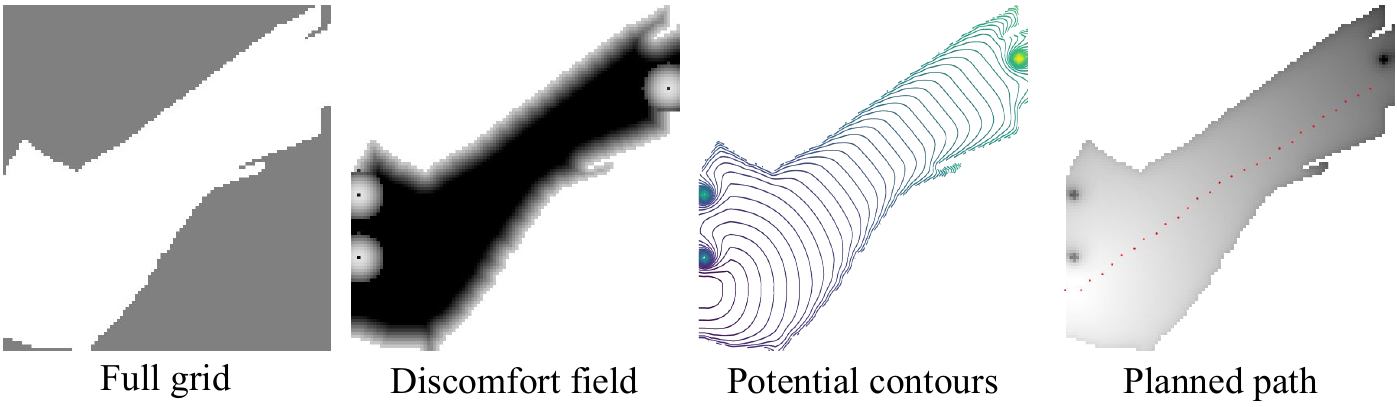}
  \caption{Example of potential field-based crowd simulation. We first project the walkable regions onto a BEV grid. Then compute a discomfort field based on scene layout and other pedestrian's positions. Finally we perform potential descent based on the potential field (planning a path perpendicular to the potential contours at each grid location.)}
  \label{fig:simulation}

\end{figure}

\subsubsection{Pedestrian Simulation}
We base our pedestrian simulation algorithm on a simplified version of the potential field algorithm in \cite{treuille2006continuum}. We start with a binary BEV map $W$ marking the walkable regions in the scene, with an additional binary obstacle map $O$ marking the position of obstacles on the street (poles, flowers, etc). We build a \textit{discomfort field} $G$ such that people prefer to be at point $x$ rather than $x'$ if $G(x) > G(x')$. The choice of $G$ is flexible; we design it to discourage people from walking too close to (1) the edge of the sidewalk region, (2) any obstaclkes on the walkable regions, or (3) other pedestrians in the scene. Finally, we compute a dynamic speed map $V$ with each grid cell set inversely proportional to the local crowd density. $G, V$ are re-computed at each time-step.

For a single pedestrian with non-obstacle start and target points $o, d \in \mathbb{R}^2$, we would like to predict a path $P = \{(x_i, y_i)\}_n$ between them in $W$. To make the path natural, we follow \cite{treuille2006continuum} to assume that the pedestrians would like to minimize:

\begin{itemize}
    \item The length of the path.
    \item The amount of time to the destination.
    \item The discomfort felt per unit time along the path.
\end{itemize}

As in \cite{treuille2006continuum}, we compute the optimal path $P$ by 
\begin{equation}
    \operatorname*{argmin}_P {\underbrace{\alpha \int_P 1 d s}_{\text {Path Length }}+\underbrace{\beta \int_P 1 d t}_{\text {Time }}+\underbrace{\gamma \int_P g d t}_{\text {Discomfort }}}
\end{equation}

where $\alpha, \beta, \gamma$ are weight parameters that could be set manually. $g$ is the discomfort value at each grid. The two variables $ds, dt$ indicate whether the integral is taken with respect to space or time, and satisfy the relationship $d s=v d t$ where $v$ is the speed. This can be further simplified to 
\begin{equation}
\operatorname*{argmin}_P \int_P C d s, \quad \text { where } \quad C \equiv \frac{\alpha v+\beta+\gamma g}{v}
\end{equation}
We start with building the \textit{unit cost field} $C$ as a weighted combination of inverse speed and discomfort. Then we use the fast marching algorithm \cite{tsitsiklis1995fastmarching} to compute a potential field $\phi$ such that $\phi=0$ at target point and otherwise satisfies the eikonal equation:
\begin{equation}
    \|\nabla \phi(\mathbf{x})\|=C
\end{equation}

At each time step, we compute the velocity of each pedestrian at position $\mathbf{x}$ and update its position with a pre-defined step size $\Delta t$:
\begin{equation}
\mathbf{x}=\mathbf{x}+V(\mathbf{x}) \frac{\nabla \phi(\mathbf{x})}{\|\nabla \phi(\mathbf{x})\|} \Delta t
\end{equation}
See Algorithm \ref{alg:simulation} for the complete procedure, and the supplementary material for implementation details. 

\begin{algorithm}
    \caption{Pedestrian Simulation Algorithm}
    \label{alg:simulation}
    \begin{algorithmic}
        
    \STATE Compute the BEV walkable map $W$, obstacle map $O$
    \STATE Initialize the discomfort map $G$, speed map $V$
    
    \FOR{each time step $\Delta t$}
    \IF{add new pedestrian} 
    \STATE{Pick origin and destination $(o, d)$ pair from the pool}
    \STATE{Initialize this pedestrian at start point $s$}
    \ENDIF{}
    \STATE Update $G, V$
    \STATE Compute the unit cost field $C$
    \STATE Construct the potential field $\phi$
    
    \FOR{each pedestrian }
    \STATE Update position with $\nabla \phi$
    \IF{distance to destination $d < \epsilon$}
    \STATE{Remove person from scene}
    \ENDIF
    \ENDFOR
    \ENDFOR
    
    \end{algorithmic}

\end{algorithm}

\subsubsection{Traffic Simulation}
We extend our attention to more complex scenes where vehicles could be present in addition to pedestrians. Specifically, we add cars to the scene if there exists a large enough "driven" region identified by CLIPSeg \cite{lueddecke2022clipseg}. The cars move with a fixed speed following the detected lane. 

We additionally detect \texttt{Crosswalk} region with CLIPSeg to activate the traffic simulation module in the system. Specifically, the system takes in a dynamic binary input indicating the status of traffic light at each crosswalk. If the light is red for cars (pedestrians), the pedestrians (cars) cross the street as normal while the cars (pedestrians) would stop before the crosswalk. We naturally accelerate or decelerate a car by checking its distance to its previous car or to the crosswalk, illustrated in Algorithm \ref{alg:car-simulation}. Here $\Delta t$ is the unit time step, and $d$ is a minimum possible distance between two cars or between the car and the crosswalk.

\begin{algorithm}
    \caption{Car Simulation Algorithm}
    \label{alg:car-simulation}
    \begin{algorithmic}
        
    \STATE Construct a list with each car pointing towards the previous car.
    
    \FOR{car with speed $v$ and acceleration/deceleration $a$}
        \STATE Compute $d_\text{car}$, the signed distance to previous car
        \STATE Compute $d_\text{cross}$, the signed distance to the crosswalk
        \IF{$0 < d_\text{cross} < v^2 / 2a + d$ and red light}
        \STATE{Decelerate,  $v \gets v - a \Delta t$ }
        \ELSIF{$0 < d_\text{car} < v^2 / 2a + d$}
        \STATE{Decelerate,  $v \gets v - a \Delta t$ }
        \ELSIF{$v < v_\text{max}$ and green light}
        \STATE{Accelerate,  $v \gets v + a \Delta t$ }
        \ENDIF
    \ENDFOR
    \end{algorithmic}
\end{algorithm}

\subsection{Rendering} 
\label{sec:rendering}
In this final stage, we render 3D pedestrians and cars into the reconstructed scene with plausible shading, shadows, and occlusions.
Our approach operates frame-by-frame; $f$ is the current video frame.

\subsubsection{Shadow Rendering}

\begin{figure}[h]
  \centering
  \includegraphics[width=\linewidth]{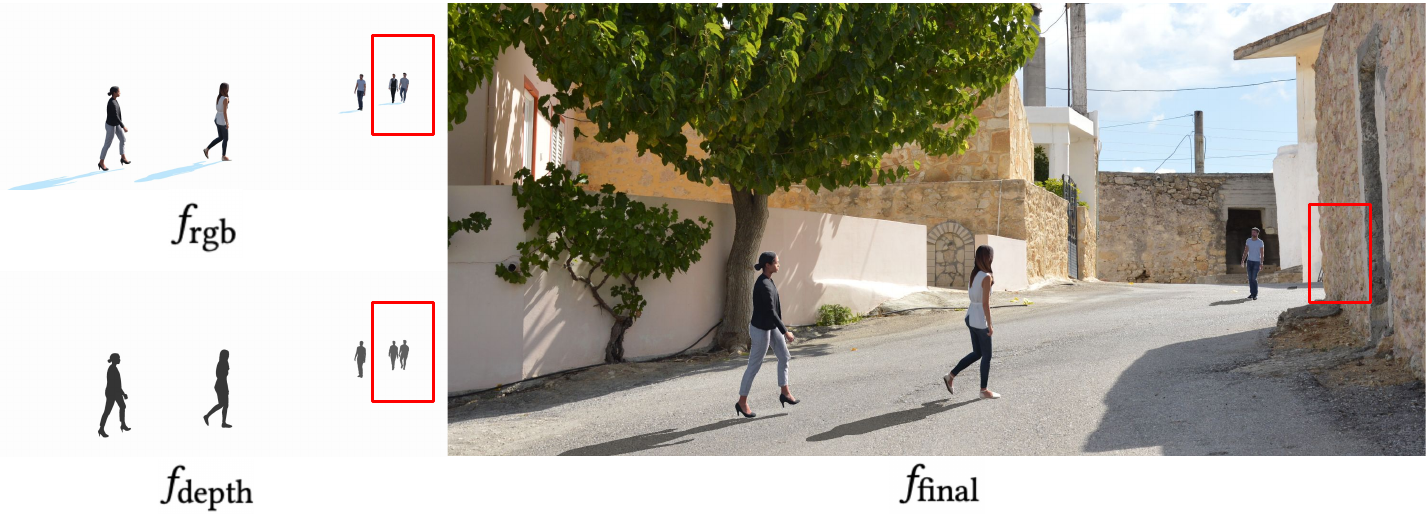}
  \caption{Shadow and occlusion rendering. We render the inserted objects with the synthesized shadow effects. Red boxes highlight occluded objects. }
  \label{fig:shadow}

\end{figure}

We start by generating shadows, shadow masks, and object masks, and compositing over the background image $B$.  The colors and depth map for $B$ are not used in this step.  Specifically, we render the scene with two different layers. First, we render a color image $f_{\text{rgb}}$ with an opaque ground plane against a pure white background, using the estimated lighting direction to cast shadows. Second, we take a depth image $f_{\text{depth}}$ with a transparent ground plane, which gives us a mask of objects in the scene without shadows (Figure \ref{fig:shadow}). 

The shadow mask $M_\text{s}$ and objects mask $M_\text{o}$ are computed by:
\begin{equation}
    M_\text{s} = \mathbb{I}[f_{\text{rgb}} > 0] - \mathbb{I}[f_{\text{depth}} < \infty]
\quad\mbox{and}\quad
    M_\text{o} = \mathbb{I}[f_{\text{depth}} < \infty]
\end{equation}
where $\mathbb{I}$ is a per-pixel indicator operator that evaluates to 0 or 1 for argument that is false or true, respectively.

We composite the final frame with shadow $f_\text{ws}$, darkening the shadow region by a  shadow color factor $s$ as follows: 
\begin{equation}
\label{eq:shadow}
    f_{\text{ws}} = f_{\text{rgb}} \odot M_\text{o} + f_{\text{rgb}} \odot M_\text{s} \times s + B \odot (1 - M_\text{s} - M_\text{o}) 
\end{equation}
where $\odot$ is element-wise multiplication.

We implement three additional refinements to improve shadow quality. First, we apply shadow color matching by taking the average color of a local non-shadowed patch and a local shadowed-patch in $B$, and compute their per-channel ratio as the shadow color factor $s$. Second, we exclude previously detected shadow regions in $B$ from $M_\text{s}$ to avoid double shadows and apply Gaussian blur to smooth out the sudden change in shadow factor. Third, we reconstruct vertical building walls if they appear in the segmentation map, and compute shadow masks cast on them in the same way as ground shadows. For cloudy days we set up a top-down light perpendicular to the ground, and apply Gaussian blur on the shadow mask to create a soft, diffuse shadow effect. See Figure \ref{fig:shadow} for an example result.

\subsubsection{Occlusion Rendering}
We use the estimated depth map with standard Z-buffering to model occlusions.
Observing that monocular depth estimation is inaccurate for thin objects, we refine their estimated depth $D_{\text{bg}}$ by computing their intersection with the ground, and assigning depth based on the estimated ground plane. We also refine $D_{\text{bg}}$ for ground pixels to fit the ground plane. 

Given the depth map $f_\text{depth}$ of rendered assets, refined background depth $D_{\text{bg}}$, $f_\text{shadow}$ from Equation \ref{eq:shadow}, and the original background scene $B$ , we composite the final result image as 
\begin{equation}
    f_{\text{final}} = \mathbb{I}[f_\text{depth} \leq D_{\text{bg}}] \odot f_{\text{ws}} + \mathbb{I}[f_\text{depth} > D_{\text{bg}}] \odot B
\end{equation}
\section{Results}
\label{sec:result}
\subsection{Data and Implementation}
We apply our system on a collection of 156 street scene images, with resolution ranging from 2160 $\times$ 1620 to 4032 $\times$ 3024. 126 of the images are taken with the main camera of iPhone 12 mini (5.76 $\times$ 4.32mm sensor) while the rest are photos from the Internet (Creative Commons or licensed). 

The inference part of the system (pre-trained segmentation and depth estimation) runs on two GeForce RTX 2080 GPUs. We use the Unity 2022 3D High Definition Rendering Pipeline (HDRP) \cite{Juliani2018UnityAG} for rendering.

\subsection{Reconstruction}
\subsubsection{Sun Estimation}
We train the sun estimation network using the dataset from \cite{chang2018panorama} containing 19093 panorama images from Google Street View as well as the ground truth sun position derived from metadata. We take images from each panorama with various perspective camera angles and FOVs and compute the corresponding sun direction as ground truth. 

We compare our system with \cite{HoldGeoffroy2019DeepSM} and \cite{Ma2016FindYW} by adding a fully connected layer to their method to predict the elevation angle. Since there is no official implementation for both papers nor released dataset, we implement them from scratch and train with our prepared datasets. On average over the test set, our azimuth prediction has an angular error of $38.7^{\circ}$ vs. $59.3^{\circ}$ \cite{HoldGeoffroy2019DeepSM} vs. $58.7^{\circ}$  \cite{Ma2016FindYW}, and our elevation prediction has an angular error of $12.3^{\circ}$ vs. $15.9^{\circ}$ \cite{HoldGeoffroy2019DeepSM} vs. $16.9^{\circ}$  \cite{Ma2016FindYW}. 

\subsubsection{Shadow Occluders}
\begin{figure}[h]
  \centering
  \includegraphics[width=\linewidth]{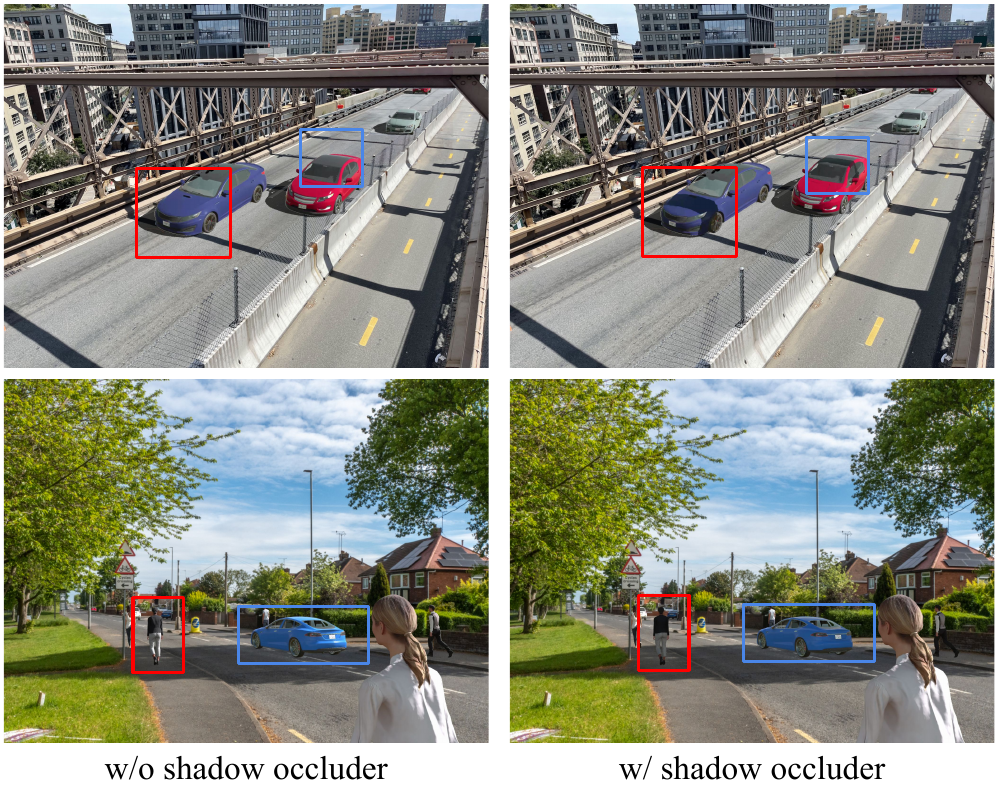}
  \caption{Rendered video frames with and without shadow occluders. See the darkened effect when an object moves into a shaded region. }
  \label{fig:occluder-result}
\end{figure}

Figure \ref{fig:occluder-result} compares the system with and without shadow occluders. Notice how the people and cars are darkened by cast shadows when they move into a shaded region. 

\subsection{Crowd Simulation}
\subsubsection{Path Planning}

\begin{figure}[h]
  \centering
  \includegraphics[width=\linewidth]{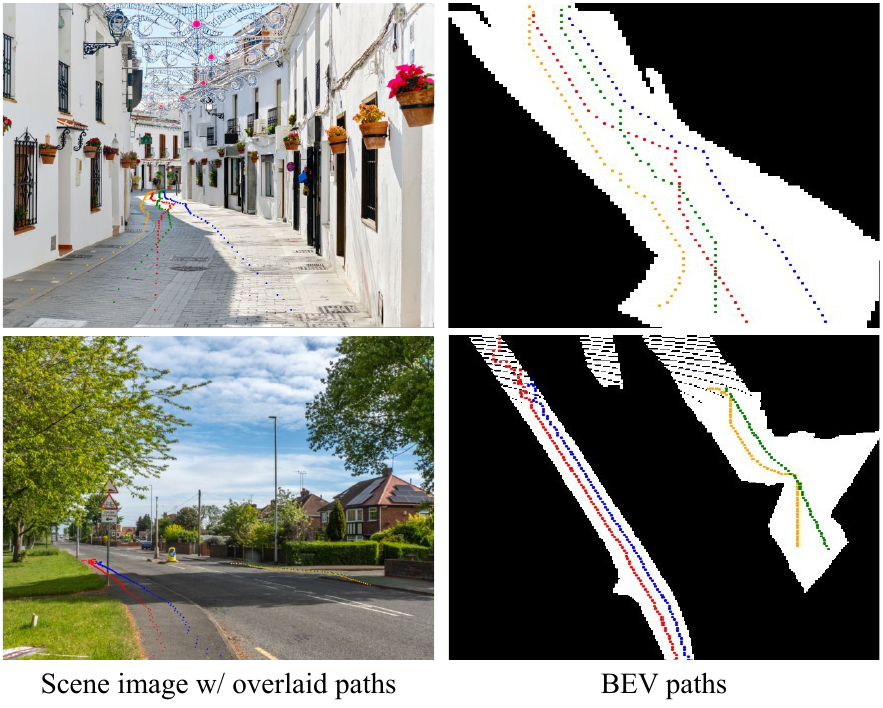}
  \caption{Crowd simulation algorithm applied to scene images. We simulate the paths in BEV space, and project the path back to RGB image space. Each color represents a pedestrian path. }
  \label{fig:simulation-result}
\end{figure}

Our system takes advantage of potential-field based gradient descent to achieve real time path planning and obstacle avoidance. We smooth the hard turns in the path in Figure \ref{fig:simulation-result} through a direction rotation function based on Slerp when we place character assets into the scene and make them follow the path. Refer to the attached video for the smooth path following behavior. 

\subsubsection{Traffic Simulation}
\begin{figure}[h]
  \centering
  \includegraphics[width=\linewidth]{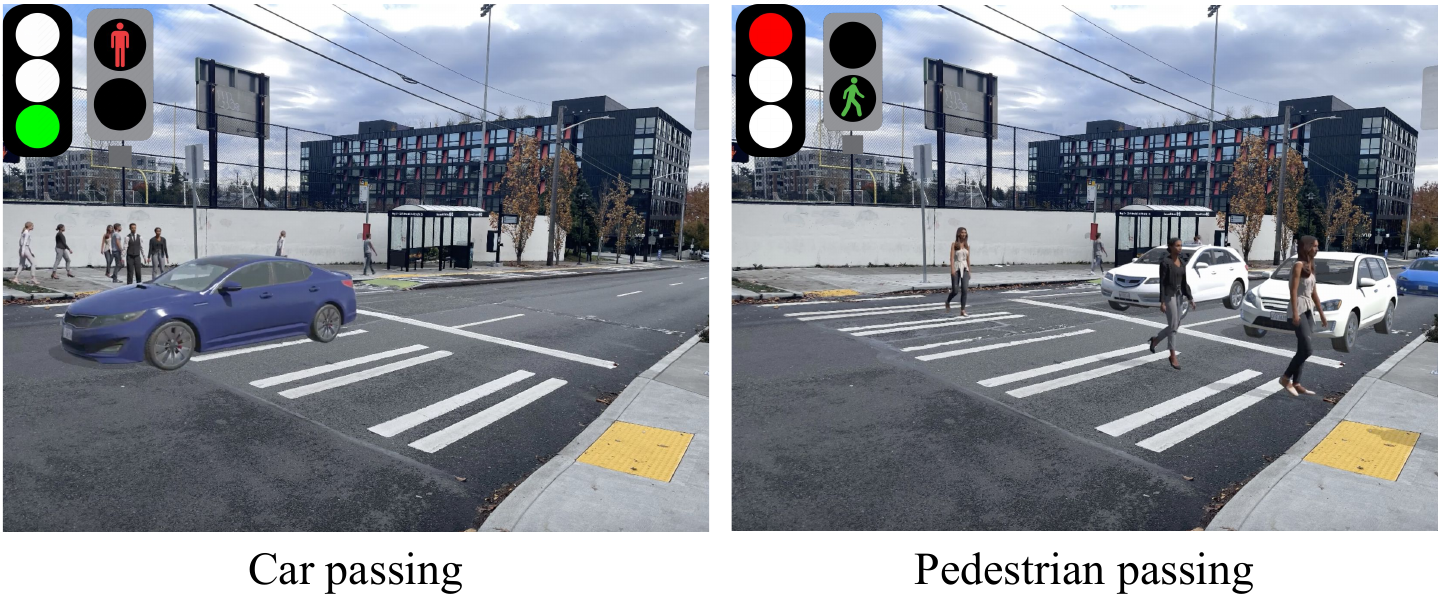}
  \caption{Example of traffic simulation. Pedestrians and cars stop before the crosswalk region when the corresponding traffic light is red.}
  \label{fig:traffic}

\end{figure}

We control the state of each crosswalk to simulate a traffic-light interaction model where pedestrians and cars wait in front of the crosswalk when the light is red. Figure \ref{fig:traffic} shows an example of an intersection. When the traffic light is green for cars and red for pedestrians, the pedestrians form a small group on the sidewalk waiting for vehicles to pass. When it's green for pedestrians and red for cars, the cars slow to a stop, while the pedestrians traverse the crosswalk. Refer to the video for a demonstration of traffic simulation. 

\subsection{Rendering}
\subsubsection{Shadows}
\begin{figure*}[h]
  \centering
  \includegraphics[width=0.96\textwidth]{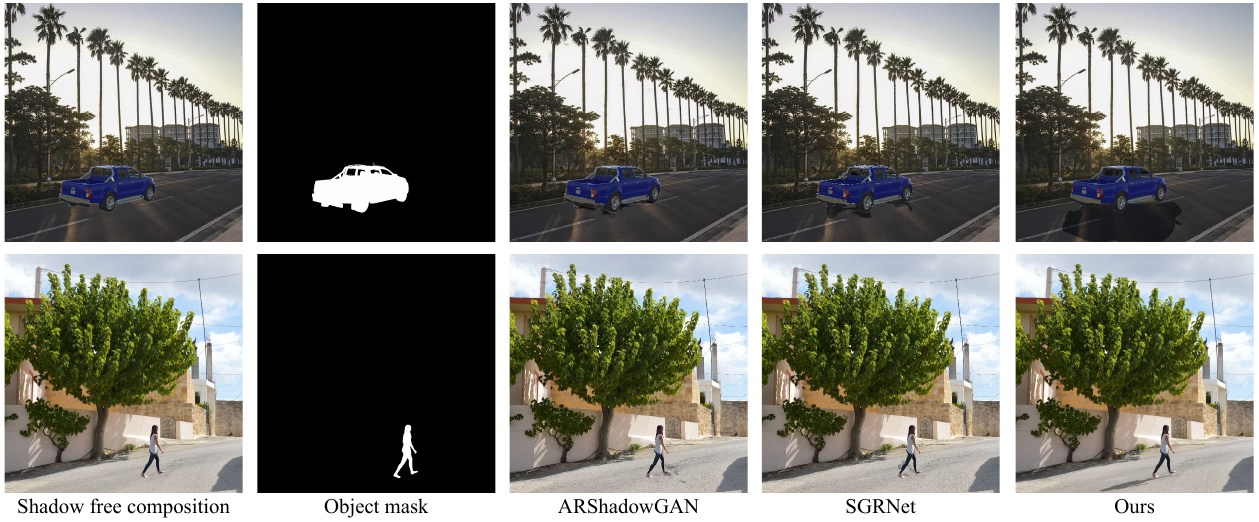}
  \caption{Comparison with recent shadow synthesis approaches ARShadowGAN \cite{liu2020mask} and SGRNet \cite{hong2021shadow}. Baseline methods cannot identify the correct shadow direction, color or generate hard shadow with the right shape.}
  \label{fig:shadow-result}

\end{figure*}

\begin{figure}[h]
  \centering
  \includegraphics[width=0.95\linewidth]{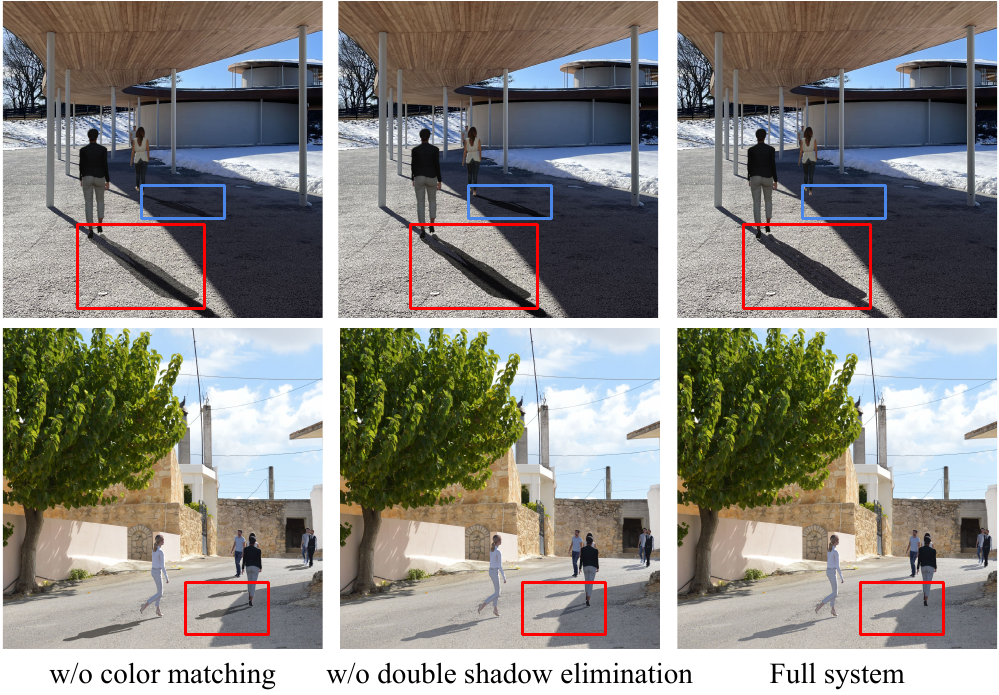}
  \caption{Adding the shadow color matching and double shadow elimination mechanism helps harmonizing existing shadows in the image with synthesized shadow. Notice how the shadow colors are unnatural without those algorithms, and are consistent/harmonized with the full system.}
  \label{fig:shadow-result-2}

\end{figure}

Because we have 3D assets, we simply render shadows with Unity's graphical rendering pipeline, yielding superior results to purely image-based methods.
Figure \ref{fig:shadow-result} compares the synthesized shadow with two state-of-the-art shadow synthesis methods \cite{liu2020mask} and \cite{hong2021shadow}. We also show our system's ability to match shadow color and handle double shadow artifacts in Figure \ref{fig:shadow-result-2}.

\subsubsection{Occlusion}
\begin{figure}[!h]
  \centering
  \includegraphics[width=0.95\linewidth]{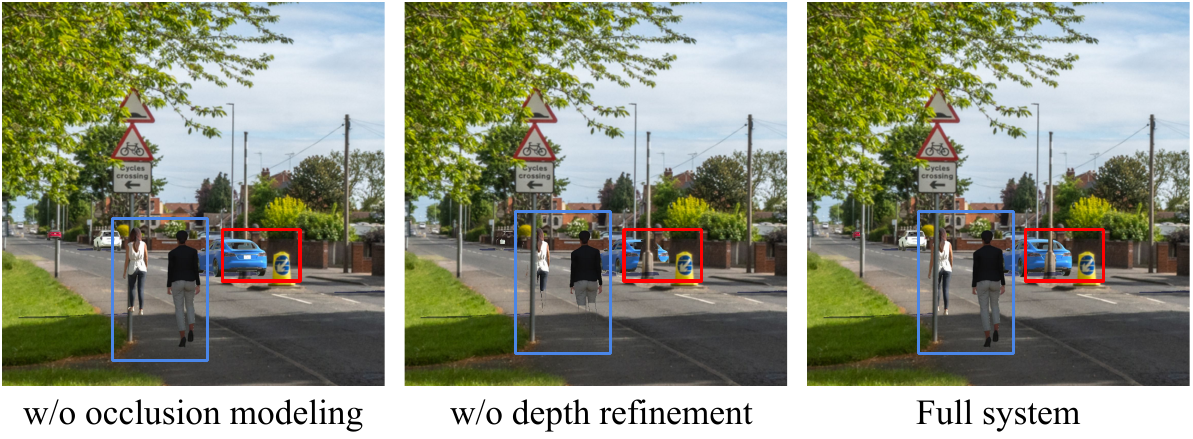}
  \caption{Monocular depth estimation could be off in shadow regions (red box) or have vague boundaries (blue box). Semantic segmentation information helps refine the depth value on grounded, thin objects.}
  \label{fig:occlusion-result}

\end{figure}

We illustrate how taking advantage of semantic segmentation information can help generate occlusion effects on thin and small objects. Refer to Figure \ref{fig:occlusion-result} for visual results. 

\subsection{Video Generation}
\begin{figure*}[!h]
  \centering
  \includegraphics[width=\linewidth]{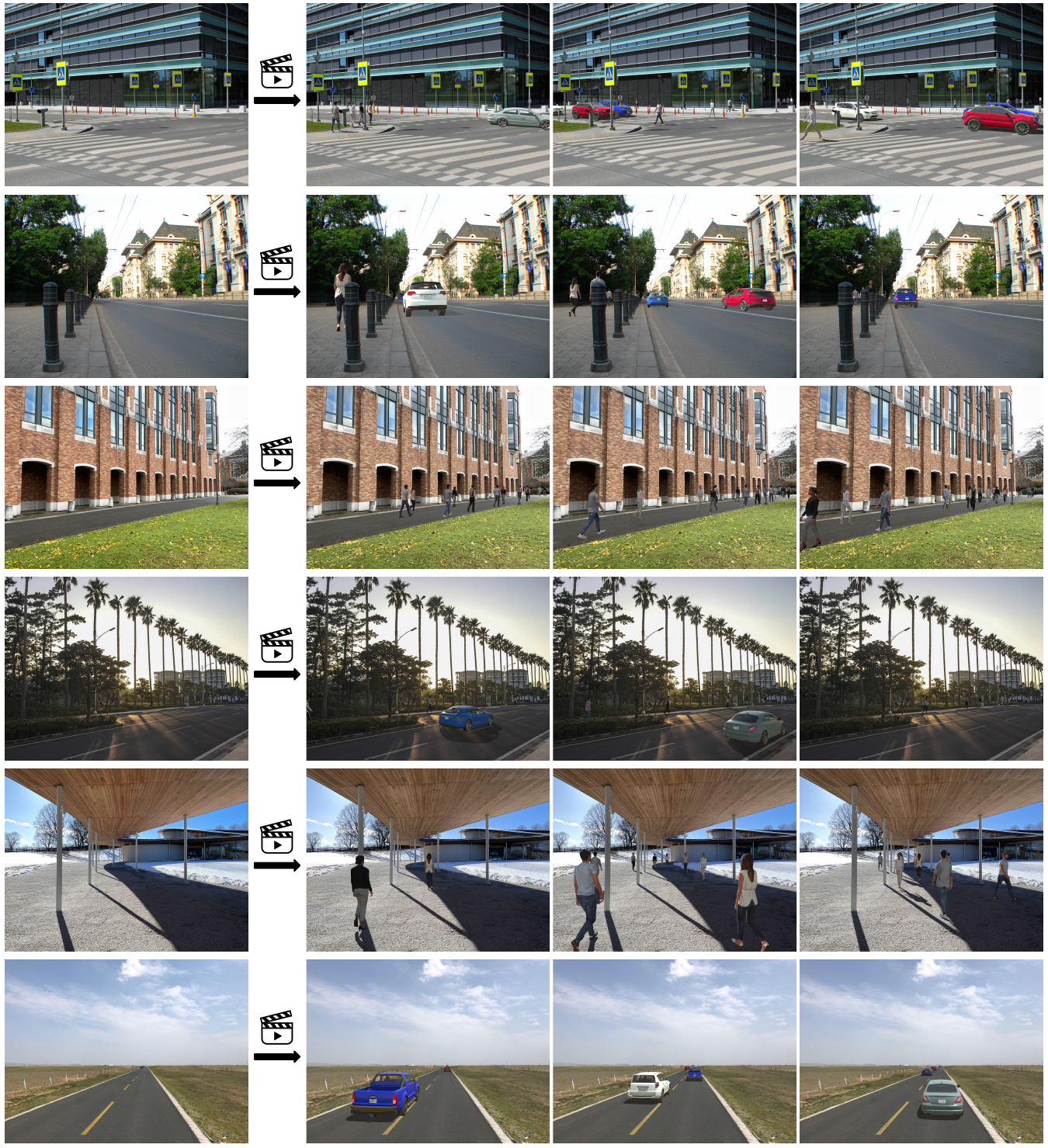}
  \caption{Visual results of the system on regular images. 
  Three representative frames are shown here for each video. Better viewed in video format in the supplementary material.}
  \label{fig:result}
\end{figure*}

We demonstrate 12 high-resolution videos results for regular images (Figure \ref{fig:result}) and one result for a panorama (Figure \ref{fig:panorama}). These results are best viewed in the supplementary video.

We record the run time of our video generation system. Reconstruction takes 5-10 second per image depending on the resolution. Simulation runs at 30 fps. Rendering is an offline process of around 10 minutes for a 1 minute video at the highest resolution (4032 $\times$ 3024) but could be made in real-time with standard shader techniques or by producing lower resolution alternatives.

\subsection{Failure Cases}

As the first end-to-end system for the task of animating street view images, we did not have a complete baseline to compare to directly. The relevant works either take different input formats than us (multi-view images or videos), or produce different output format (static images). Instead, here We count the failure cases at each step. Note that different categories' counts may have overlapping. For example, one single scene may suffer from both depth estimation issue and semantic segmentation issue. The statistics of failure cases on our dataset of 156 images are as follows:

\begin{itemize}
    \item Depth estimation: 38 (wrong scale/occlusion)
    \item Semantic segmentation: 45 (failure to identify complex road structure)
    \item Sun estimation: 19 (wrong intensity/direction)
    \item Shadow detection: 15 (fail to detect shadow region)
    \item Crosswalk detection: 8 (fail to identify crosswalk)
    \item Crowd simulation: 10 (noticeable collision)
    \item Shadow rendering: 16 (wrong shadow color matching/double shadow)
    \item Inpainting: 13 (unrealistic inpainting)
    
\end{itemize}

Segmentation and depth estimation models contribute to most of our failure cases, and our results will improve with progress on these topics. Our contribution is to build a composable system and using algorithms proven to be state-of-the-art in the literature.


\section{Conclusion and Discussion}
\label{sec:conclusion}

In this paper, we present a system that automatically populates a still scene image with naturally behaving pedestrians and vehicles. We describe the algorithms for each component of the system (reconstruction, simulation, rendering) as a combination of state-of-the-art deep learning methods and traditional simulation/rendering. We illustrate the quality of our approach with rendered videos. 

\begin{figure}[!h]
  \centering
  \includegraphics[width=\linewidth]{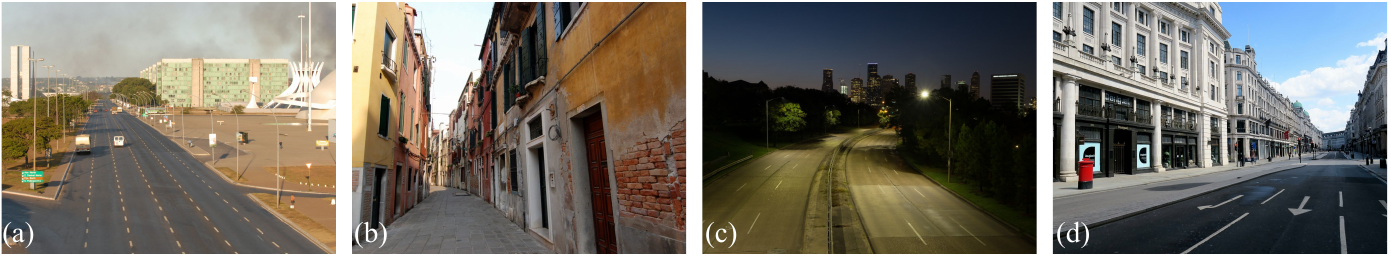}
  \caption{Failure cases: (a) out-of-distribution camera angle. (b) ambiguous shadow region. (c) curved lanes. (d) insufficent cues for sun direction.}
  \label{fig:limitations}

\end{figure}

\textbf{Limitations:} Our system does not currently handle curved lanes or hills/slopes, complete shadows cast on inserted objects when the full shadow is not in camera view, automatic traffic direction detection, recovering the full extent of obstacles (incl. back half not seen), and casting shadows onto arbitrary scene geometry; and our lighting model is approximate (not a full global illumination rendering).  As our system relies on the accuracy of pre-trained models (depth, segmentation, shadow), it fails on out of distribution images, e.g., with non-street-level camera angles and difficult lighting conditions. Indeed, we are presenting a method that doesn't attempt to improve scene understanding, but takes advantage of existing information in an integrated way to achieve appealing visual effects. With advances in pre-trained scene understanding models, our system can be adapted to a wider range of scenes.  See Figure \ref{fig:limitations} for a few examples of scenes we don't currently handle.

\textbf{Diversity and Ethical Considerations:} We acknowledge that our paper does not explicitly model any population distribution so it could potentially generate videos that misrepresent a place's public space, and tolerated outfits, vehicle types, etc. The assets we use to generate visualizations in the paper and demo video show limited diversity in appearance, gender, and ethnicity. However, our system can be applied to any type of assets compatible with the game engine. It remains worth exploring problem to model the actual population distribution based on a scene layout.


\begin{acks}
This work was supported by the UW Reality Lab, Meta, Google, OPPO, and Amazon.
\end{acks}

\bibliographystyle{ACM-Reference-Format}
\bibliography{reference}


\begin{thebibliography}{57}


\ifx \showCODEN    \undefined \def \showCODEN     #1{\unskip}     \fi
\ifx \showDOI      \undefined \def \showDOI       #1{#1}\fi
\ifx \showISBNx    \undefined \def \showISBNx     #1{\unskip}     \fi
\ifx \showISBNxiii \undefined \def \showISBNxiii  #1{\unskip}     \fi
\ifx \showISSN     \undefined \def \showISSN      #1{\unskip}     \fi
\ifx \showLCCN     \undefined \def \showLCCN      #1{\unskip}     \fi
\ifx \shownote     \undefined \def \shownote      #1{#1}          \fi
\ifx \showarticletitle \undefined \def \showarticletitle #1{#1}   \fi
\ifx \showURL      \undefined \def \showURL       {\relax}        \fi
\providecommand\bibfield[2]{#2}
\providecommand\bibinfo[2]{#2}
\providecommand\natexlab[1]{#1}
\providecommand\showeprint[2][]{arXiv:#2}

\bibitem[Azadi et~al\mbox{.}(2020)]%
        {azadi2018compositional}
\bibfield{author}{\bibinfo{person}{Samaneh Azadi}, \bibinfo{person}{Deepak
  Pathak}, \bibinfo{person}{Sayna Ebrahimi}, {and} \bibinfo{person}{Trevor
  Darrell}.} \bibinfo{year}{2020}\natexlab{}.
\newblock \showarticletitle{Compositional GAN: Learning Image-conditional
  Binary Composition}.
\newblock \bibinfo{journal}{\emph{International Journal of Computer Vision
  (IJCV)}}  \bibinfo{volume}{128} (\bibinfo{year}{2020}),
  \bibinfo{pages}{2570--2585}.
\newblock


\bibitem[Bhat et~al\mbox{.}(2021)]%
        {sfaroop2021adabins}
\bibfield{author}{\bibinfo{person}{S.~Farooq Bhat}, \bibinfo{person}{I.
  Alhashim}, {and} \bibinfo{person}{P. Wonka}.}
  \bibinfo{year}{2021}\natexlab{}.
\newblock \showarticletitle{AdaBins: Depth Estimation Using Adaptive Bins}. In
  \bibinfo{booktitle}{\emph{Proceedings of the IEEE/CVF Conference on Computer
  Vision and Pattern Recognition (CVPR)}}. \bibinfo{pages}{4008--4017}.
\newblock


\bibitem[Chang et~al\mbox{.}(2018)]%
        {chang2018panorama}
\bibfield{author}{\bibinfo{person}{Shih-Hsiu Chang}, \bibinfo{person}{Ching-Ya
  Chiu}, \bibinfo{person}{Chia-Sheng Chang}, \bibinfo{person}{Kuo-Wei Chen},
  \bibinfo{person}{Chih-Yuan Yao}, \bibinfo{person}{Ruen-Rone Lee}, {and}
  \bibinfo{person}{Hung-Kuo Chu}.} \bibinfo{year}{2018}\natexlab{}.
\newblock \showarticletitle{Generating 360 Outdoor Panorama Dataset with
  Reliable Sun Position Estimation}. In \bibinfo{booktitle}{\emph{SIGGRAPH Asia
  Posters}} (Tokyo, Japan). \bibinfo{publisher}{Association for Computing
  Machinery}, \bibinfo{address}{New York, NY, USA}, Article
  \bibinfo{articleno}{22}, \bibinfo{numpages}{2}~pages.
\newblock
\showISBNx{9781450360630}


\bibitem[Chen et~al\mbox{.}(2021)]%
        {chen2021geosim}
\bibfield{author}{\bibinfo{person}{Yun Chen}, \bibinfo{person}{Frieda Rong},
  \bibinfo{person}{Shivam Duggal}, \bibinfo{person}{Shenlong Wang},
  \bibinfo{person}{Xinchen Yan}, \bibinfo{person}{Sivabalan Manivasagam},
  \bibinfo{person}{Shangjie Xue}, \bibinfo{person}{Ersin Yumer}, {and}
  \bibinfo{person}{Raquel Urtasun}.} \bibinfo{year}{2021}\natexlab{}.
\newblock \showarticletitle{GeoSim: Realistic Video Simulation via
  Geometry-Aware Composition for Self-Driving}. In
  \bibinfo{booktitle}{\emph{Proceedings of the IEEE/CVF Conference on Computer
  Vision and Pattern Recognition (CVPR)}}.
\newblock


\bibitem[Chien et~al\mbox{.}(2017)]%
        {chien2017nonexistent}
\bibfield{author}{\bibinfo{person}{Jui-Ting Chien}, \bibinfo{person}{Chia-Jung
  Chou}, \bibinfo{person}{Ding-Jie Chen}, {and} \bibinfo{person}{Hwann-Tzong
  Chen}.} \bibinfo{year}{2017}\natexlab{}.
\newblock \showarticletitle{Detecting Nonexistent Pedestrians}. In
  \bibinfo{booktitle}{\emph{IEEE/CVF Conference on Computer Vision and Pattern
  Recognition Workshop (CVPRW)}}.
\newblock


\bibitem[Cordts et~al\mbox{.}(2016)]%
        {Cordts2016Cityscapes}
\bibfield{author}{\bibinfo{person}{Marius Cordts}, \bibinfo{person}{Mohamed
  Omran}, \bibinfo{person}{Sebastian Ramos}, \bibinfo{person}{Timo Rehfeld},
  \bibinfo{person}{Markus Enzweiler}, \bibinfo{person}{Rodrigo Benenson},
  \bibinfo{person}{Uwe Franke}, \bibinfo{person}{Stefan Roth}, {and}
  \bibinfo{person}{Bernt Schiele}.} \bibinfo{year}{2016}\natexlab{}.
\newblock \showarticletitle{The Cityscapes Dataset for Semantic Urban Scene
  Understanding}. In \bibinfo{booktitle}{\emph{Proceedings of the IEEE/CVF
  Conference on Computer Vision and Pattern Recognition (CVPR)}}.
\newblock


\bibitem[Davtyan and Favaro(2022)]%
        {davtyan2022cvg}
\bibfield{author}{\bibinfo{person}{Aram Davtyan} {and} \bibinfo{person}{Paolo
  Favaro}.} \bibinfo{year}{2022}\natexlab{}.
\newblock \showarticletitle{Controllable Video Generation Through Global and
  Local Motion Dynamics}. In \bibinfo{booktitle}{\emph{Proceedings of the
  European Conference on Computer Vision (ECCV)}}. \bibinfo{pages}{68--84}.
\newblock


\bibitem[Epstein et~al\mbox{.}(2022)]%
        {epstein2022blobgan}
\bibfield{author}{\bibinfo{person}{Dave Epstein}, \bibinfo{person}{Taesung
  Park}, \bibinfo{person}{Richard Zhang}, \bibinfo{person}{Eli Shechtman},
  {and} \bibinfo{person}{Alexei~A. Efros}.} \bibinfo{year}{2022}\natexlab{}.
\newblock \showarticletitle{BlobGAN: Spatially Disentangled Scene
  Representations}.
\newblock \bibinfo{journal}{\emph{European Conference on Computer Vision
  (ECCV)}} (\bibinfo{year}{2022}).
\newblock


\bibitem[He et~al\mbox{.}(2015)]%
        {He2015DeepRL}
\bibfield{author}{\bibinfo{person}{Kaiming He}, \bibinfo{person}{X. Zhang},
  \bibinfo{person}{Shaoqing Ren}, {and} \bibinfo{person}{Jian Sun}.}
  \bibinfo{year}{2015}\natexlab{}.
\newblock \showarticletitle{Deep Residual Learning for Image Recognition}. In
  \bibinfo{booktitle}{\emph{Proceedings of the IEEE/CVF Conference on Computer
  Vision and Pattern Recognition (CVPR)}}. \bibinfo{pages}{770--778}.
\newblock


\bibitem[Hold-Geoffroy et~al\mbox{.}(2019)]%
        {HoldGeoffroy2019DeepSM}
\bibfield{author}{\bibinfo{person}{Yannick Hold-Geoffroy},
  \bibinfo{person}{Akshaya Athawale}, {and} \bibinfo{person}{Jean-François
  Lalonde}.} \bibinfo{year}{2019}\natexlab{}.
\newblock \showarticletitle{Deep Sky Modeling for Single Image Outdoor Lighting
  Estimation}. In \bibinfo{booktitle}{\emph{Proceedings of IEEE/CVF Conference
  on Computer Vision and Pattern Recognition (CVPR)}}.
  \bibinfo{pages}{6920--6928}.
\newblock


\bibitem[Holynski et~al\mbox{.}(2021)]%
        {holynski2021animating}
\bibfield{author}{\bibinfo{person}{Aleksander Holynski},
  \bibinfo{person}{Brian~L. Curless}, \bibinfo{person}{Steven~M. Seitz}, {and}
  \bibinfo{person}{Richard Szeliski}.} \bibinfo{year}{2021}\natexlab{}.
\newblock \showarticletitle{Animating Pictures With Eulerian Motion Fields}. In
  \bibinfo{booktitle}{\emph{Proceedings of the IEEE/CVF Conference on Computer
  Vision and Pattern Recognition (CVPR)}}. \bibinfo{pages}{5810--5819}.
\newblock


\bibitem[Hong et~al\mbox{.}(2022)]%
        {hong2021shadow}
\bibfield{author}{\bibinfo{person}{Yan Hong}, \bibinfo{person}{Li Niu}, {and}
  \bibinfo{person}{Jianfu Zhang}.} \bibinfo{year}{2022}\natexlab{}.
\newblock \showarticletitle{Shadow Generation for Composite Image in Real-world
  Scenes}. In \bibinfo{booktitle}{\emph{Proceedings of the AAAI Conference on
  Artificial Intelligence (AAAI)}}.
\newblock


\bibitem[Hu et~al\mbox{.}(2022)]%
        {hu2022makeitmove}
\bibfield{author}{\bibinfo{person}{Yaosi Hu}, \bibinfo{person}{Chong Luo},
  {and} \bibinfo{person}{Zhenzhong Chen}.} \bibinfo{year}{2022}\natexlab{}.
\newblock \showarticletitle{Make It Move: Controllable Image-to-Video
  Generation With Text Descriptions}. In \bibinfo{booktitle}{\emph{Proceedings
  of the IEEE/CVF Conference on Computer Vision and Pattern Recognition
  (CVPR)}}. \bibinfo{pages}{18219--18228}.
\newblock


\bibitem[Huang et~al\mbox{.}(2022a)]%
        {huang2022elicit}
\bibfield{author}{\bibinfo{person}{Yangyi Huang}, \bibinfo{person}{Hongwei Yi},
  \bibinfo{person}{Weiyang Liu}, \bibinfo{person}{Haofan Wang},
  \bibinfo{person}{Boxi Wu}, \bibinfo{person}{Wenxiao Wang},
  \bibinfo{person}{Binbin Lin}, \bibinfo{person}{Debing Zhang}, {and}
  \bibinfo{person}{Deng Cai}.} \bibinfo{year}{2022}\natexlab{a}.
\newblock \bibinfo{title}{One-shot Implicit Animatable Avatars with Model-based
  Priors}.
\newblock
\newblock
\showeprint[arxiv]{arXiv:2212.02469}


\bibitem[Huang et~al\mbox{.}(2022b)]%
        {huang2022physically}
\bibfield{author}{\bibinfo{person}{Ziyuan Huang}, \bibinfo{person}{Zhengping
  Zhou}, \bibinfo{person}{Yung-Yu Chuang}, \bibinfo{person}{Jiajun Wu}, {and}
  \bibinfo{person}{C.~Karen Liu}.} \bibinfo{year}{2022}\natexlab{b}.
\newblock \bibinfo{title}{Physically Plausible Animation of Human Upper Body
  from a Single Image}.
\newblock
\newblock
\showeprint[arxiv]{2212.04741}


\bibitem[Juliani et~al\mbox{.}(2018)]%
        {Juliani2018UnityAG}
\bibfield{author}{\bibinfo{person}{Arthur Juliani},
  \bibinfo{person}{Vincent-Pierre Berges}, \bibinfo{person}{Esh Vckay},
  \bibinfo{person}{Yuan Gao}, \bibinfo{person}{Hunter Henry},
  \bibinfo{person}{Marwan Mattar}, {and} \bibinfo{person}{Danny Lange}.}
  \bibinfo{year}{2018}\natexlab{}.
\newblock \bibinfo{title}{Unity: A General Platform for Intelligent Agents}.
\newblock
\newblock
\showeprint[arxiv]{1809.02627}


\bibitem[Karras et~al\mbox{.}(2023)]%
        {karras2023dreampose}
\bibfield{author}{\bibinfo{person}{Johanna Karras}, \bibinfo{person}{Aleksander
  Holynski}, \bibinfo{person}{Ting-Chun Wang}, {and} \bibinfo{person}{Ira
  Kemelmacher-Shlizerman}.} \bibinfo{year}{2023}\natexlab{}.
\newblock \bibinfo{title}{DreamPose: Fashion Image-to-Video Synthesis via
  Stable Diffusion}.
\newblock
\newblock
\showeprint[arxiv]{arXiv:2304.06025}


\bibitem[Karsch et~al\mbox{.}(2011)]%
        {Karsch2011insertion}
\bibfield{author}{\bibinfo{person}{Kevin Karsch}, \bibinfo{person}{Varsha
  Hedau}, \bibinfo{person}{David Forsyth}, {and} \bibinfo{person}{Derek
  Hoiem}.} \bibinfo{year}{2011}\natexlab{}.
\newblock \showarticletitle{Rendering Synthetic Objects into Legacy
  Photographs}.
\newblock \bibinfo{journal}{\emph{ACM Trans. Graph.}} \bibinfo{volume}{30},
  \bibinfo{number}{6} (\bibinfo{date}{dec} \bibinfo{year}{2011}),
  \bibinfo{pages}{1–12}.
\newblock
\showISSN{0730-0301}
\urldef\tempurl%
\url{https://doi.org/10.1145/2070781.2024191}
\showDOI{\tempurl}


\bibitem[Karsch et~al\mbox{.}(2014)]%
        {karsch2014insertion}
\bibfield{author}{\bibinfo{person}{Kevin Karsch}, \bibinfo{person}{Kalyan
  Sunkavalli}, \bibinfo{person}{Sunil Hadap}, \bibinfo{person}{Nathan Carr},
  \bibinfo{person}{Hailin Jin}, \bibinfo{person}{Rafael Fonte},
  \bibinfo{person}{Michael Sittig}, {and} \bibinfo{person}{David Forsyth}.}
  \bibinfo{year}{2014}\natexlab{}.
\newblock \showarticletitle{Automatic Scene Inference for 3D Object
  Compositing}.
\newblock \bibinfo{journal}{\emph{ACM Trans. Graph.}} \bibinfo{volume}{33},
  \bibinfo{number}{3}, Article \bibinfo{articleno}{32} (\bibinfo{date}{jun}
  \bibinfo{year}{2014}), \bibinfo{numpages}{15}~pages.
\newblock
\showISSN{0730-0301}
\urldef\tempurl%
\url{https://doi.org/10.1145/2602146}
\showDOI{\tempurl}


\bibitem[Kim et~al\mbox{.}(2021)]%
        {kim2021drivegan}
\bibfield{author}{\bibinfo{person}{Seung~Wook Kim}, \bibinfo{person}{},
  \bibinfo{person}{Jonah Philion}, \bibinfo{person}{Antonio Torralba}, {and}
  \bibinfo{person}{Sanja Fidler}.} \bibinfo{year}{2021}\natexlab{}.
\newblock \showarticletitle{{DriveGAN: Towards a Controllable High-Quality
  Neural Simulation}}. In \bibinfo{booktitle}{\emph{IEEE Conference on Computer
  Vision and Pattern Recognition (CVPR)}}.
\newblock


\bibitem[Lee et~al\mbox{.}(2018)]%
        {lee2018insert}
\bibfield{author}{\bibinfo{person}{Donghoon Lee}, \bibinfo{person}{Sifei Liu},
  \bibinfo{person}{Jinwei Gu}, \bibinfo{person}{Ming-Yu Liu},
  \bibinfo{person}{Ming-Hsuan Yang}, {and} \bibinfo{person}{Jan Kautz}.}
  \bibinfo{year}{2018}\natexlab{}.
\newblock \showarticletitle{Context-Aware Synthesis and Placement of Object
  Instances}. In \bibinfo{booktitle}{\emph{Proceedings of the International
  Conference on Neural Information Processing Systems (NeurIPS)}}.
  \bibinfo{pages}{10414–10424}.
\newblock


\bibitem[Lee et~al\mbox{.}(2019)]%
        {Lee2019InsertingVI}
\bibfield{author}{\bibinfo{person}{Donghoon Lee}, \bibinfo{person}{Tomas
  Pfister}, {and} \bibinfo{person}{Ming-Hsuan Yang}.}
  \bibinfo{year}{2019}\natexlab{}.
\newblock \showarticletitle{Inserting Videos Into Videos}. In
  \bibinfo{booktitle}{\emph{Proceedings of the IEEE/CVF Conference on Computer
  Vision and Pattern Recognition (CVPR)}}.
\newblock


\bibitem[Lin et~al\mbox{.}(2018)]%
        {lin2018stgan}
\bibfield{author}{\bibinfo{person}{Chen-Hsuan Lin}, \bibinfo{person}{Ersin
  Yumer}, \bibinfo{person}{Oliver Wang}, \bibinfo{person}{Eli Shechtman}, {and}
  \bibinfo{person}{Simon Lucey}.} \bibinfo{year}{2018}\natexlab{}.
\newblock \showarticletitle{ST-GAN: Spatial Transformer Generative Adversarial
  Networks for Image Compositing}. In \bibinfo{booktitle}{\emph{Proceedings of
  the IEEE/CVF Conference on Computer Vision and Pattern Recognition (CVPR)}}.
  \bibinfo{pages}{9455--9464}.
\newblock


\bibitem[Liu et~al\mbox{.}(2020)]%
        {liu2020mask}
\bibfield{author}{\bibinfo{person}{Daquan Liu}, \bibinfo{person}{Chengjiang
  Long}, \bibinfo{person}{Hongpan Zhang}, \bibinfo{person}{Hanning Yu},
  \bibinfo{person}{Xinzhi Dong}, {and} \bibinfo{person}{Chunxia Xiao}.}
  \bibinfo{year}{2020}\natexlab{}.
\newblock \showarticletitle{ARShadowGAN: Shadow Generative Adversarial Network
  for Augmented Reality in Single Light Scenes}. In
  \bibinfo{booktitle}{\emph{Proceedings of the IEEE/CVF Conference on Computer
  Vision and Pattern Recognition (CVPR)}}.
\newblock


\bibitem[L\"uddecke and Ecker(2022)]%
        {lueddecke2022clipseg}
\bibfield{author}{\bibinfo{person}{Timo L\"uddecke} {and}
  \bibinfo{person}{Alexander Ecker}.} \bibinfo{year}{2022}\natexlab{}.
\newblock \showarticletitle{Image Segmentation Using Text and Image Prompts}.
  In \bibinfo{booktitle}{\emph{Proceedings of the IEEE/CVF Conference on
  Computer Vision and Pattern Recognition (CVPR)}}.
  \bibinfo{pages}{7086--7096}.
\newblock


\bibitem[Ma et~al\mbox{.}(2016)]%
        {Ma2016FindYW}
\bibfield{author}{\bibinfo{person}{Wei-Chiu Ma}, \bibinfo{person}{Shenlong
  Wang}, \bibinfo{person}{Marcus~A. Brubaker}, \bibinfo{person}{Sanja Fidler},
  {and} \bibinfo{person}{Raquel Urtasun}.} \bibinfo{year}{2016}\natexlab{}.
\newblock \showarticletitle{Find your way by observing the sun and other
  semantic cues}. In \bibinfo{booktitle}{\emph{Proceedings of IEEE
  International Conference on Robotics and Automation (ICRA)}}.
  \bibinfo{pages}{6292--6299}.
\newblock


\bibitem[Ma et~al\mbox{.}(2023)]%
        {ma2023directed}
\bibfield{author}{\bibinfo{person}{Wan-Duo~Kurt Ma}, \bibinfo{person}{J.~P.
  Lewis}, \bibinfo{person}{W.~Bastiaan Kleijn}, {and} \bibinfo{person}{Thomas
  Leung}.} \bibinfo{year}{2023}\natexlab{}.
\newblock \bibinfo{title}{Directed Diffusion: Direct Control of Object
  Placement through Attention Guidance}.
\newblock
\newblock
\showeprint[arxiv]{2302.13153}


\bibitem[Mallya et~al\mbox{.}(2022)]%
        {mallya2022implicit}
\bibfield{author}{\bibinfo{person}{Arun Mallya}, \bibinfo{person}{Ting-Chun
  Wang}, {and} \bibinfo{person}{Ming-Yu Liu}.} \bibinfo{year}{2022}\natexlab{}.
\newblock \showarticletitle{{Implicit Warping for Animation with Image Sets}}.
  In \bibinfo{booktitle}{\emph{Proceedings of the International Conference on
  Neural Information Processing Systems (NeurIPS)}}.
\newblock


\bibitem[Menapace et~al\mbox{.}(2022)]%
        {menapace2022pvg}
\bibfield{author}{\bibinfo{person}{Willi Menapace}, \bibinfo{person}{St\'ephane
  Lathuili\`ere}, \bibinfo{person}{Aliaksandr Siarohin},
  \bibinfo{person}{Christian Theobalt}, \bibinfo{person}{Sergey Tulyakov},
  \bibinfo{person}{Vladislav Golyanik}, {and} \bibinfo{person}{Elisa Ricci}.}
  \bibinfo{year}{2022}\natexlab{}.
\newblock \showarticletitle{Playable Environments: Video Manipulation in Space
  and Time}. In \bibinfo{booktitle}{\emph{Proceedings of the IEEE/CVF
  Conference on Computer Vision and Pattern Recognition (CVPR)}}.
  \bibinfo{pages}{3584--3593}.
\newblock


\bibitem[Menapace et~al\mbox{.}(2021)]%
        {menapace2021pvg}
\bibfield{author}{\bibinfo{person}{Willi Menapace}, \bibinfo{person}{Stephane
  Lathuiliere}, \bibinfo{person}{Sergey Tulyakov}, \bibinfo{person}{Aliaksandr
  Siarohin}, {and} \bibinfo{person}{Elisa Ricci}.}
  \bibinfo{year}{2021}\natexlab{}.
\newblock \showarticletitle{Playable Video Generation}. In
  \bibinfo{booktitle}{\emph{Proceedings of the IEEE/CVF Conference on Computer
  Vision and Pattern Recognition (CVPR)}}. \bibinfo{pages}{10061--10070}.
\newblock


\bibitem[Nguyen-Phuoc et~al\mbox{.}(2020)]%
        {BlockGAN2020}
\bibfield{author}{\bibinfo{person}{Thu Nguyen-Phuoc},
  \bibinfo{person}{Christian Richardt}, \bibinfo{person}{Long Mai},
  \bibinfo{person}{Yong-Liang Yang}, {and} \bibinfo{person}{Niloy Mitra}.}
  \bibinfo{year}{2020}\natexlab{}.
\newblock \showarticletitle{BlockGAN: Learning 3D Object-aware Scene
  Representations from Unlabelled Images}. In
  \bibinfo{booktitle}{\emph{Advances in Neural Information Processing Systems
  33}}.
\newblock


\bibitem[Ni et~al\mbox{.}(2023)]%
        {ni2023conditional}
\bibfield{author}{\bibinfo{person}{Haomiao Ni}, \bibinfo{person}{Changhao Shi},
  \bibinfo{person}{Kai Li}, \bibinfo{person}{Sharon~X. Huang}, {and}
  \bibinfo{person}{Martin~Renqiang Min}.} \bibinfo{year}{2023}\natexlab{}.
\newblock \bibinfo{title}{Conditional Image-to-Video Generation with Latent
  Flow Diffusion Models}.
\newblock
\newblock
\showeprint[arxiv]{2303.13744}


\bibitem[Niemeyer and Geiger(2021)]%
        {niemeyer2020GIRAFFE}
\bibfield{author}{\bibinfo{person}{Michael Niemeyer} {and}
  \bibinfo{person}{Andreas Geiger}.} \bibinfo{year}{2021}\natexlab{}.
\newblock \showarticletitle{GIRAFFE: Representing Scenes as Compositional
  Generative Neural Feature Fields}. In \bibinfo{booktitle}{\emph{Proc. IEEE
  Conf. on Computer Vision and Pattern Recognition (CVPR)}}.
\newblock


\bibitem[Okabe et~al\mbox{.}(2011)]%
        {okabe2011fluid}
\bibfield{author}{\bibinfo{person}{Makoto Okabe}, \bibinfo{person}{Ken Anjyor},
  {and} \bibinfo{person}{Rikio Onai}.} \bibinfo{year}{2011}\natexlab{}.
\newblock \showarticletitle{Creating Fluid Animation from a Single Image using
  Video Database}.
\newblock \bibinfo{journal}{\emph{Computer Graphics Forum}}
  \bibinfo{volume}{30}, \bibinfo{number}{7} (\bibinfo{year}{2011}),
  \bibinfo{pages}{1973--1982}.
\newblock
\urldef\tempurl%
\url{https://doi.org/10.1111/j.1467-8659.2011.02062.x}
\showDOI{\tempurl}


\bibitem[Peng et~al\mbox{.}(2021)]%
        {peng2021animatable}
\bibfield{author}{\bibinfo{person}{Sida Peng}, \bibinfo{person}{Junting Dong},
  \bibinfo{person}{Qianqian Wang}, \bibinfo{person}{Shangzhan Zhang},
  \bibinfo{person}{Qing Shuai}, \bibinfo{person}{Xiaowei Zhou}, {and}
  \bibinfo{person}{Hujun Bao}.} \bibinfo{year}{2021}\natexlab{}.
\newblock \showarticletitle{Animatable Neural Radiance Fields for Modeling
  Dynamic Human Bodies}. In \bibinfo{booktitle}{\emph{Proceedings of the
  IEEE/CVF International Conference on Computer Vision (ICCV)}}.
  \bibinfo{pages}{14314--14323}.
\newblock


\bibitem[P{\'e}rez et~al\mbox{.}(2003)]%
        {Prez2003PoissonIE}
\bibfield{author}{\bibinfo{person}{Patrick P{\'e}rez}, \bibinfo{person}{Michel
  Gangnet}, {and} \bibinfo{person}{Andrew Blake}.}
  \bibinfo{year}{2003}\natexlab{}.
\newblock \showarticletitle{Poisson image editing}. In
  \bibinfo{booktitle}{\emph{ACM SIGGRAPH}}. \bibinfo{publisher}{Association for
  Computing Machinery}, \bibinfo{address}{New York, NY, USA},
  \bibinfo{pages}{313–318}.
\newblock


\bibitem[Pumarola et~al\mbox{.}(2019)]%
        {pumarola2019ganimation}
\bibfield{author}{\bibinfo{person}{A. Pumarola}, \bibinfo{person}{A. Agudo},
  \bibinfo{person}{A.M. Martinez}, \bibinfo{person}{A. Sanfeliu}, {and}
  \bibinfo{person}{F. Moreno-Noguer}.} \bibinfo{year}{2019}\natexlab{}.
\newblock \showarticletitle{GANimation: One-Shot Anatomically Consistent Facial
  Animation}.
\newblock \bibinfo{journal}{\emph{International Journal of Computer Vision
  (IJCV)}} (\bibinfo{year}{2019}).
\newblock


\bibitem[Rombach et~al\mbox{.}(2022)]%
        {Rombach2022SD}
\bibfield{author}{\bibinfo{person}{Robin Rombach}, \bibinfo{person}{Andreas
  Blattmann}, \bibinfo{person}{Dominik Lorenz}, \bibinfo{person}{Patrick
  Esser}, {and} \bibinfo{person}{Bj\"orn Ommer}.}
  \bibinfo{year}{2022}\natexlab{}.
\newblock \showarticletitle{High-Resolution Image Synthesis With Latent
  Diffusion Models}. In \bibinfo{booktitle}{\emph{Proceedings of the IEEE/CVF
  Conference on Computer Vision and Pattern Recognition (CVPR)}}.
  \bibinfo{pages}{10684--10695}.
\newblock


\bibitem[Song et~al\mbox{.}(2022)]%
        {song2022objectstitch}
\bibfield{author}{\bibinfo{person}{Yizhi Song}, \bibinfo{person}{Zhifei Zhang},
  \bibinfo{person}{Zhe Lin}, \bibinfo{person}{Scott Cohen},
  \bibinfo{person}{Brian Price}, \bibinfo{person}{Jianming Zhang},
  \bibinfo{person}{Soo~Ye Kim}, {and} \bibinfo{person}{Daniel Aliaga}.}
  \bibinfo{year}{2022}\natexlab{}.
\newblock \bibinfo{title}{ObjectStitch: Generative Object Compositing}.
\newblock
\newblock
\showeprint[arxiv]{2212.00932}


\bibitem[Sun et~al\mbox{.}(2020)]%
        {hiddenfootprints2020eccv}
\bibfield{author}{\bibinfo{person}{Jin Sun}, \bibinfo{person}{Hadar
  Averbuch-Elor}, \bibinfo{person}{Qianqian Wang}, {and} \bibinfo{person}{Noah
  Snavely}.} \bibinfo{year}{2020}\natexlab{}.
\newblock \showarticletitle{Hidden Footprints: Learning Contextual Walkability
  from 3D Human Trails}. In \bibinfo{booktitle}{\emph{Proceedings of the
  European Conference on Computer Vision (ECCV)}}.
\newblock


\bibitem[Tang et~al\mbox{.}(2022)]%
        {tang2022lighting}
\bibfield{author}{\bibinfo{person}{Jiajun Tang}, \bibinfo{person}{Yongjie Zhu},
  \bibinfo{person}{Haoyu Wang}, \bibinfo{person}{Jun~Hoong Chan},
  \bibinfo{person}{Si Li}, {and} \bibinfo{person}{Boxin Shi}.}
  \bibinfo{year}{2022}\natexlab{}.
\newblock \showarticletitle{Estimating Spatially-Varying Lighting In Urban
  Scenes With Disentangled Representation}. In
  \bibinfo{booktitle}{\emph{Proceedings of the European Conference on Computer
  Vision (ECCV)}}. \bibinfo{publisher}{Springer Nature Switzerland},
  \bibinfo{address}{Cham}, \bibinfo{pages}{454--469}.
\newblock


\bibitem[Treuille et~al\mbox{.}(2006)]%
        {treuille2006continuum}
\bibfield{author}{\bibinfo{person}{Adrien Treuille}, \bibinfo{person}{Seth
  Cooper}, {and} \bibinfo{person}{Zoran Popovi\'{c}}.}
  \bibinfo{year}{2006}\natexlab{}.
\newblock \showarticletitle{Continuum Crowds}.
\newblock \bibinfo{journal}{\emph{ACM Transactions on Graphics}}
  \bibinfo{volume}{25}, \bibinfo{number}{3} (\bibinfo{date}{july}
  \bibinfo{year}{2006}), \bibinfo{pages}{1160–1168}.
\newblock
\showISSN{0730-0301}
\urldef\tempurl%
\url{https://doi.org/10.1145/1141911.1142008}
\showDOI{\tempurl}


\bibitem[Tsitsiklis(1995)]%
        {tsitsiklis1995fastmarching}
\bibfield{author}{\bibinfo{person}{J.N. Tsitsiklis}.}
  \bibinfo{year}{1995}\natexlab{}.
\newblock \showarticletitle{Efficient Algorithms for Globally Optimal
  Trajectories}.
\newblock \bibinfo{journal}{\emph{IEEE Trans. Automat. Control}}
  \bibinfo{volume}{40}, \bibinfo{number}{9} (\bibinfo{year}{1995}),
  \bibinfo{pages}{1528--1538}.
\newblock


\bibitem[Wang et~al\mbox{.}(2019)]%
        {wang2019fewshotvid2vid}
\bibfield{author}{\bibinfo{person}{Ting-Chun Wang}, \bibinfo{person}{Ming-Yu
  Liu}, \bibinfo{person}{Andrew Tao}, \bibinfo{person}{Guilin Liu},
  \bibinfo{person}{Jan Kautz}, {and} \bibinfo{person}{Bryan Catanzaro}.}
  \bibinfo{year}{2019}\natexlab{}.
\newblock \showarticletitle{Few-shot Video-to-Video Synthesis}. In
  \bibinfo{booktitle}{\emph{Proceedings of the International Conference on
  Neural Information Processing Systems (NeurIPS)}}.
\newblock


\bibitem[Wang et~al\mbox{.}(2020)]%
        {wang20people}
\bibfield{author}{\bibinfo{person}{Yifan Wang}, \bibinfo{person}{Brian~L
  Curless}, {and} \bibinfo{person}{Steven~M Seitz}.}
  \bibinfo{year}{2020}\natexlab{}.
\newblock \showarticletitle{People as Scene Probes}. In
  \bibinfo{booktitle}{\emph{Proceedings of the European Conference on Computer
  Vision (ECCV)}}.
\newblock


\bibitem[Wang et~al\mbox{.}(2021)]%
        {wang2021repopulating}
\bibfield{author}{\bibinfo{person}{Yifan Wang}, \bibinfo{person}{Andrew Liu},
  \bibinfo{person}{Richard Tucker}, \bibinfo{person}{Jiajun Wu},
  \bibinfo{person}{Brian~L Curless}, \bibinfo{person}{Steven~M Seitz}, {and}
  \bibinfo{person}{Noah Snavely}.} \bibinfo{year}{2021}\natexlab{}.
\newblock \showarticletitle{Repopulating Street Scenes}. In
  \bibinfo{booktitle}{\emph{Proceedings of the IEEE/CVF Conference on Computer
  Vision and Pattern Recognition (CVPR)}}.
\newblock


\bibitem[Wang et~al\mbox{.}(2022)]%
        {wang2022neural}
\bibfield{author}{\bibinfo{person}{Zian Wang}, \bibinfo{person}{Wenzheng Chen},
  \bibinfo{person}{David Acuna}, \bibinfo{person}{Jan Kautz}, {and}
  \bibinfo{person}{Sanja Fidler}.} \bibinfo{year}{2022}\natexlab{}.
\newblock \showarticletitle{Neural Light Field Estimation for Street Scenes
  with Differentiable Virtual Object Insertion}. In
  \bibinfo{booktitle}{\emph{Proceedings of the European Conference on Computer
  Vision (ECCV)}}. \bibinfo{publisher}{Springer Nature Switzerland},
  \bibinfo{address}{Cham}, \bibinfo{pages}{380--397}.
\newblock


\bibitem[Weng et~al\mbox{.}(2019)]%
        {weng2019wakeup}
\bibfield{author}{\bibinfo{person}{Chung-Yi Weng}, \bibinfo{person}{Brian
  Curless}, {and} \bibinfo{person}{Ira Kemelmacher}.}
  \bibinfo{year}{2019}\natexlab{}.
\newblock \showarticletitle{Photo Wake-Up: 3D Character Animation From a Single
  Photo}. In \bibinfo{booktitle}{\emph{Proceedings of the IEEE/CVF Conference
  on Computer Vision and Pattern Recognition (CVPR)}}.
  \bibinfo{pages}{5901--5910}.
\newblock


\bibitem[Xie et~al\mbox{.}(2021)]%
        {xie2021segformer}
\bibfield{author}{\bibinfo{person}{Enze Xie}, \bibinfo{person}{Wenhai Wang},
  \bibinfo{person}{Zhiding Yu}, \bibinfo{person}{Anima Anandkumar},
  \bibinfo{person}{Jose~M Alvarez}, {and} \bibinfo{person}{Ping Luo}.}
  \bibinfo{year}{2021}\natexlab{}.
\newblock \showarticletitle{SegFormer: Simple and Efficient Design for Semantic
  Segmentation with Transformers}. In \bibinfo{booktitle}{\emph{Proceedings of
  Neural Information Processing Systems (NeurIPS)}}.
\newblock


\bibitem[Xu et~al\mbox{.}(2022)]%
        {xu2022discoscene}
\bibfield{author}{\bibinfo{person}{Yinghao Xu}, \bibinfo{person}{Menglei Chai},
  \bibinfo{person}{Zifan Shi}, \bibinfo{person}{Sida Peng},
  \bibinfo{person}{Skorokhodov Ivan}, \bibinfo{person}{Siarohin Aliaksandr},
  \bibinfo{person}{Ceyuan Yang}, \bibinfo{person}{Yujun Shen},
  \bibinfo{person}{Hsin-Ying Lee}, \bibinfo{person}{Bolei Zhou}, {and}
  \bibinfo{person}{Tulyakov Sergy}.} \bibinfo{year}{2022}\natexlab{}.
\newblock \bibinfo{title}{DiscoScene: Spatially Disentangled Generative
  Radiance Field for Controllable 3D-aware Scene Synthesis}.
\newblock
\newblock
\showeprint[arxiv]{2212.11984}


\bibitem[Xue et~al\mbox{.}(2022)]%
        {xue2022giraffehd}
\bibfield{author}{\bibinfo{person}{Yang Xue}, \bibinfo{person}{Yuheng Li},
  \bibinfo{person}{Krishna~Kumar Singh}, {and} \bibinfo{person}{Yong~Jae Lee}.}
  \bibinfo{year}{2022}\natexlab{}.
\newblock \showarticletitle{GIRAFFE HD: A High-Resolution 3D-aware Generative
  Model}. In \bibinfo{booktitle}{\emph{CVPR}}.
\newblock


\bibitem[Yoon et~al\mbox{.}(2021)]%
        {yoon2021poseguided}
\bibfield{author}{\bibinfo{person}{Jae~Shin Yoon}, \bibinfo{person}{Lingjie
  Liu}, \bibinfo{person}{Vladislav Golyanik}, \bibinfo{person}{Kripasindhu
  Sarkar}, \bibinfo{person}{Hyun~Soo Park}, {and} \bibinfo{person}{Christian
  Theobalt}.} \bibinfo{year}{2021}\natexlab{}.
\newblock \showarticletitle{Pose-Guided Human Animation from a Single Image in
  the Wild}. In \bibinfo{booktitle}{\emph{Proceedings of the IEEE/CVF
  Conference on Computer Vision and Pattern Recognition (CVPR)}}.
\newblock


\bibitem[Yu et~al\mbox{.}(2022)]%
        {yu2022semantic}
\bibfield{author}{\bibinfo{person}{Wei Yu}, \bibinfo{person}{Wenxin Chen},
  \bibinfo{person}{Songheng Yin}, \bibinfo{person}{Steve Easterbrook}, {and}
  \bibinfo{person}{Animesh Garg}.} \bibinfo{year}{2022}\natexlab{}.
\newblock \showarticletitle{Modular Action Concept Grounding in Semantic Video
  Prediction}. In \bibinfo{booktitle}{\emph{Proceedings of the IEEE/CVF
  Conference on Computer Vision and Pattern Recognition (CVPR)}}.
\newblock


\bibitem[Zhan et~al\mbox{.}(2019)]%
        {zhan2019sfgan}
\bibfield{author}{\bibinfo{person}{Fangneng Zhan}, \bibinfo{person}{Hongyuan
  Zhu}, {and} \bibinfo{person}{Shijian Lu}.} \bibinfo{year}{2019}\natexlab{}.
\newblock \showarticletitle{Spatial Fusion GAN for Image Synthesis}. In
  \bibinfo{booktitle}{\emph{Proceedings of the IEEE/CVF Conference on Computer
  Vision and Pattern Recognition (CVPR)}}.
\newblock


\bibitem[Zhang et~al\mbox{.}(2021)]%
        {zhang2021vid2player}
\bibfield{author}{\bibinfo{person}{Haotian Zhang}, \bibinfo{person}{Cristobal
  Sciutto}, \bibinfo{person}{Maneesh Agrawala}, {and} \bibinfo{person}{Kayvon
  Fatahalian}.} \bibinfo{year}{2021}\natexlab{}.
\newblock \showarticletitle{Vid2player: Controllable video sprites that behave
  and appear like professional tennis players}.
\newblock \bibinfo{journal}{\emph{ACM Transactions on Graphics (TOG)}}
  \bibinfo{volume}{40}, \bibinfo{number}{3} (\bibinfo{year}{2021}),
  \bibinfo{pages}{1--16}.
\newblock


\bibitem[Zhang et~al\mbox{.}(2023)]%
        {zhang2023vid2player3d}
\bibfield{author}{\bibinfo{person}{Haotian Zhang}, \bibinfo{person}{Ye Yuan},
  \bibinfo{person}{Viktor Makoviychuk}, \bibinfo{person}{Yunrong Guo},
  \bibinfo{person}{Sanja Fidler}, \bibinfo{person}{Xue~Bin Peng}, {and}
  \bibinfo{person}{Kayvon Fatahalian}.} \bibinfo{year}{2023}\natexlab{}.
\newblock \showarticletitle{Learning Physically Simulated Tennis Skills from
  Broadcast Videos}.
\newblock \bibinfo{journal}{\emph{ACM Transactions on Graphics (TOG)}}
  (\bibinfo{year}{2023}), \bibinfo{pages}{1--16}.
\newblock


\bibitem[Zhu et~al\mbox{.}(2021)]%
        {zhu2021shadow}
\bibfield{author}{\bibinfo{person}{Lei Zhu}, \bibinfo{person}{Ke Xu},
  \bibinfo{person}{Zhanghan Ke}, {and} \bibinfo{person}{Rynson~W.H. Lau}.}
  \bibinfo{year}{2021}\natexlab{}.
\newblock \showarticletitle{Mitigating Intensity Bias in Shadow Detection via
  Feature Decomposition and Reweighting}. In
  \bibinfo{booktitle}{\emph{Proceedings of IEEE/CVF International Conference on
  Computer Vision (ICCV)}}. \bibinfo{pages}{4682--4691}.
\newblock


\end{thebibliography}

\begin{figure}[!ht]
  \centering
  \includegraphics[width=0.95\linewidth]{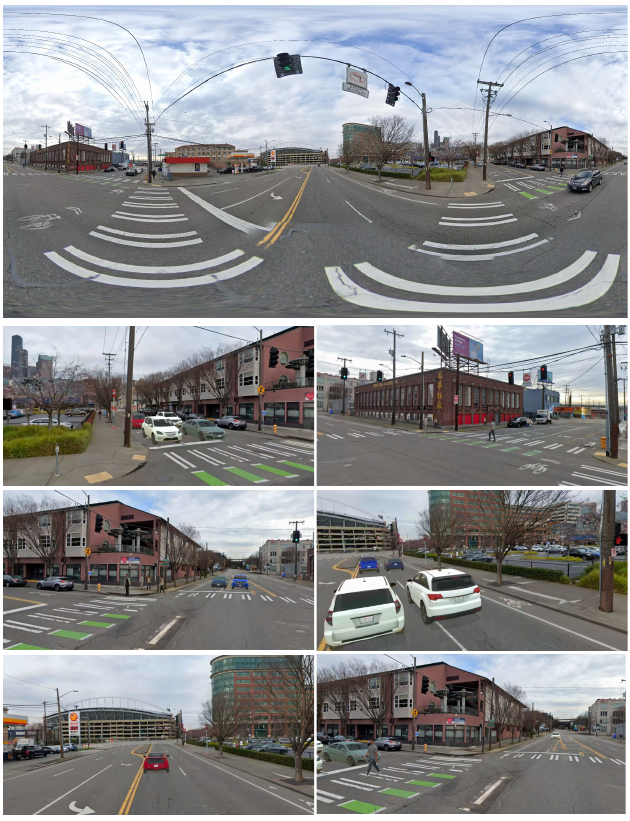}
  \caption{Visual results of the system on panorama.  Top: original panorama. Down: captured still image from different angles. Better viewed in video format in the supplementary material.}
  \label{fig:panorama}
\end{figure}

\clearpage
\end{document}